\documentclass{article}

\usepackage{microtype}
\usepackage{graphicx}
\usepackage{subcaption}
\usepackage{booktabs} 

\usepackage{hyperref}

\usepackage[preprint]{icml2026}

\usepackage{amsmath}
\usepackage{amssymb}
\usepackage{mathtools}
\usepackage{amsthm}
\usepackage{bbm}
\newcommand{\tool}{a_{\text{tool}}}
\newcommand{\mem}{a_{\text{mem}}}

\usepackage[capitalize,noabbrev]{cleveref}

\theoremstyle{plain}
\newtheorem{theorem}{Theorem}[section]
\newtheorem{proposition}[theorem]{Proposition}

\theoremstyle{definition}
\newtheorem{definition}[theorem]{Definition}

\theoremstyle{remark}

\usepackage{multirow}
\usepackage{multicol}
\usepackage{colortbl} 
\definecolor{lightblue}{RGB}{192,224,255} 
\newcommand{\up}[1]{\textsubscript{\color{green!70!black}{↑#1}}}
\newcommand{\down}[1]{\textsubscript{\color{red!70!black}{↓#1}}}
\usepackage[textsize=tiny]{todonotes}

\begin{document}

\twocolumn[
  \icmltitle{The Tool-Overuse Illusion: Why Does LLM Prefer External Tools over Internal Knowledge?}



  \icmlsetsymbol{equal}{*}

  \begin{icmlauthorlist}
    \icmlauthor{Yirong Zeng}{yyy,comp}
    \icmlauthor{Shen You}{yyy,comp}
    \icmlauthor{Yufei Liu}{sch,comp}
    \icmlauthor{Qunyao Du}{yyy}
    \icmlauthor{Xiao Ding}{yyy}
    \icmlauthor{Yutai Hou}{comp} \\
    \icmlauthor{Yuxian Wang}{comp}
    \icmlauthor{Wu Ning}{comp}
    \icmlauthor{Haonan Song}{comp}
    \icmlauthor{Dandan Tu}{comp}
    \icmlauthor{Bibo Cai}{yyy}
    \icmlauthor{Ting Liu}{yyy}

  \end{icmlauthorlist}

  \icmlaffiliation{yyy}{Harbin Institute of Technology, SCIR}
  \icmlaffiliation{sch}{Peking University}
  \icmlaffiliation{comp}{Huawei Technologies Co., Ltd}

  \icmlcorrespondingauthor{Xiao Ding}{xding@ir.hit.edu.cn}
  \icmlcorrespondingauthor{Yutai Hou}{houyutai@huawei.com}

  \icmlkeywords{Machine Learning, ICML}

  \vskip 0.3in
]



\printAffiliationsAndNotice{}  

\begin{abstract}
Equipping LLMs with external tools effectively addresses internal reasoning limitations. 
However, it introduces a critical yet under-explored phenomenon: tool overuse, the unnecessary tool-use during reasoning.
In this paper, we first reveal this phenomenon is pervasive across diverse LLMs. 
We then experimentally elucidate its underlying mechanisms through two key lenses:
(1) First, by analyzing tool-use behavior across different internal knowledge availability regions, 
we identify a \textit{knowledge epistemic illusion}: models misjudge internal knowledge boundaries and fail to accurately perceive their actual knowledge availability. 
To mitigate this, we propose a knowledge-aware epistemic boundary alignment strategy based on direct preference optimization, which reduces tool usage in by 82.8\% while yielding an accuracy improvement.
(2) Second, we establish a causal link between reward structures and tool-use behavior by visualizing the tool-augmented training process.
It reveals that \textit{outcome-only rewards} inadvertently encourage tool overuse by rewarding only final correctness, regardless of tool efficiency. 
To verify this, we balance reward signals during training rather than relying on outcome-only rewards, cutting unnecessary tool calls by 66.7\% (7B) and 60.7\% (32B) without sacrificing accuracy.
Finally, we provide theoretical justification in this two lenses to understand tool overuse.
\end{abstract}

\begin{figure}[th]
  \begin{center}
    \centerline{\includegraphics[width=0.95\columnwidth]{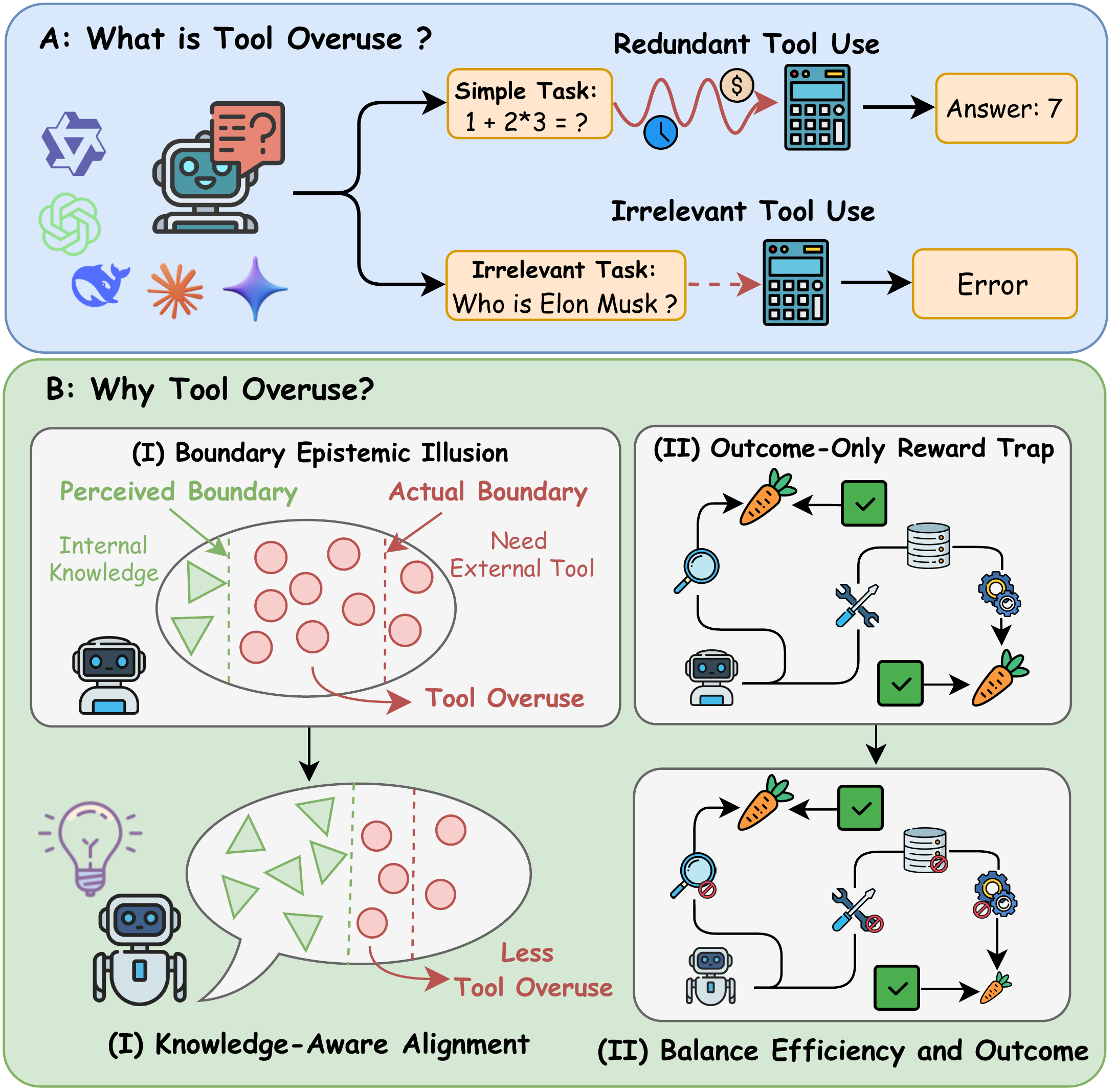}}
    \caption{
        An overview of tool overuse in LLMs. 
        We provide a systematic investigation into why models invoke external tools unnecessarily. 
        Our analysis reveals that this widespread phenomenon is driven by two primary mechanisms: knowledge epistemic illusion (a miscalibration of internal knowledge) and outcome-only rewards (optimization biases during training).
    }
    \label{fig:intro}
  \end{center}
  \vskip -0.3in
\end{figure}

\section{Introduction}
Large Language Models (LLMs) can significantly improve their reasoning capabilities by interacting with external tools, a paradigm known as tool-integrated reasoning (TIR) \citep{xue2025simpletir,feng2025retool}.
It represents a promising frontier to transcend their internal knowledge boundaries \citep{chen2024unlocking,chen20252},
and solve problems that are infeasible through text-only reasoning.
However, TIR also raises a critical problem, i.e., LLMs often overuse tools in problem-solving.

Tool overuse refers to calling external tools in unnecessary context,
such as calling irrelevant tools or executing actions that the model’s parametric knowledge could resolve internally.
Such behaviors directly incur avoidable resource consumption, and yield no information gain \citep{qian2025smart}. 
Recent research extensively employs tools to enhance complex problem-solving; however, the underlying mechanisms and systematic impacts of tool overuse have not be unexplored.

In this paper, we present the first investigation of the tool overuse problem and elucidate its underlying mechanisms (see \cref{fig:intro}).
We begin by quantifying the prevalence of this phenomenon across diverse models (e.g., frontier, OSS and RLVR-trained\footnote{Reinforcement Learning with Verifiable Rewards, a mainstream train approach to enhancing tool-integrated reasoning.} LLMs), finding an average of 0.93 unnecessary tool calls per query that requires no external assistance.
Crucially, enabling tool use leads to a $3.29\sim14.48\%$ drop in accuracy on questions solvable using internal knowledge alone.
Finally, through the following two analyses, we demystify the mechanisms driving this behavior and offer actionable insights for mitigation:

Firstly, we measure internal knowledge availability via $avg@1024$ (i.e., the average accuracy in 1,024 independently inference per sample) and analyze tool-use behavior across knowledge regimes.
Our analysis reveals knowledge epistemic illusion, where models fail to perceive the actual internal knowledge availability.
Moreover, we observe higher internal knowledge availability does not lead to fewer tool calls.
For instance, Qwen3-8B averages 2.2 tool calls per query even in high-knowledge regions (e.g., $avg@1024 > 0.8$). 
They effectively hallucinate having reached the boundary of their internal knowledge.
To bridge this gap, we construct preference pairs that yield correct answers with minimal against excessive tool usage.
By employing knowledge-aware direct preference optimization, we align the model's perceived boundaries with its actual capabilities. 
This strategy mitigates the tool-overuse problem of 32B-scale model by 82.8\% while yielding a 3\% improvement in response accuracy.

Secondly, recent research primarily focused on applying RLVR to enhance the tool-use capabilities of LLMs \citep{lin2025understanding}, which successfully improve task accuracy via outcome-driven rewards \citep{bai2025towards,mai2025agent,xue2025simpletir}.
However, our analysis reveals that outcome-only rewards act as a misleading incentive, inadvertently reinforcing the model's tool overuse illusion. 
Specifically, we observe that the tool overuse problem worsens progressively as training steps increase, ultimately reaching a 65\% increase in tool call turns.
To support this observation experimentally, we propose a reward strategy that balances outcome correctness and tool efficiency by perceiving model's capability ceiling through large inter-group sampling size during rollouts.
Results on mathematical reasoning benchmarks demonstrate that our strategy reduces tool-call turns by 66.7\% for the 7B model and 60.7\% for the 32B model, while maintaining the response accuracy.

Besides, we provide a theoretical proof to substantiate these two mechanisms, offering a comprehensive framework for understanding and mitigating tool overuse in LLMs.
Overall, this study provides a deep understanding of tool overuse, yielding mechanistic insights for building reliable tool-augmented LLMs.

\section{Preliminaries}
\subsection{Problem Formulation}
We formulate the tool-integrated reasoning process as a partially observable markov decision process, denoted as a tuple $\langle \mathcal{S}, \mathcal{A}, P, R, \Omega, O \rangle$. 
At each time step $t$, the agent receives an observation history $h_t$ and selects an action $a_t$ from the action space $\mathcal{A}$.

\noindent\textbf{Hybrid Action Space.} Distinct from standard language modeling, the action space $\mathcal{A} = \mathcal{A}_{\text{gen}} \cup \mathcal{A}_{\text{tool}}$ comprises two disjoint sets:
\begin{itemize}
    \item \textbf{Generation Actions ($\mathcal{A}_{\text{gen}}$):} Tokens from the vocabulary $\mathcal{V}$ used for internal reasoning or generating final responses.
    \item \textbf{Tool Actions ($\mathcal{A}_{\text{tool}}$):} A set of predefined APIs. An action $a_t \in \mathcal{A}_{\text{tool}}$ involves generating a structured call $c_t = (\text{func\_name}, \text{args})$.
\end{itemize}

\textbf{Tool Overuse}. Tool overuse refers to the phenomenon where LLMs perform tool actions $a_t \in \mathcal{A}_\text{tool}$ unnecessarily within the decision-making process, specifically when the internal parametric knowledge is sufficient to derive the correct answer, or when the tool call provides no information gain.


\textbf{Taxonomy}.
We formally define two modes of overuse that hinder efficient reasoning:
(1) {Redundant usage}, 
the model delegates trivial tasks to external tools despite possessing sufficient parametric knowledge. 
This behavior introduces unnecessary latency and dependency.
(2) {Irrelevant usage},
the model hallucinates tool capabilities to compensate for reasoning gaps, invoking irrelevant tools. This yields no informational gain and further increases context burden.

\subsection{RLVR Algorithms}
Reinforcement Learning with Verifiable Rewards (RLVR) has established itself as a paradigm for endowing LLMs with robust tool-integrated reasoning capabilities. 
By optimizing against outcome-oriented signals, RLVR enables models to explore and consolidate complex decision-making paths that are often missed in supervised learning.
In this work, we adopt the {DAPO} algorithm \citep{yu2025dapo}, 
a state-of-the-art RL framework designed for large-scale reasoning tasks. 
DAPO enhances the standard Group Relative Policy Optimization (GRPO \citep{guo2025deepseek}) by introducing four key techniques, e.g., \textit{Decoupled Clipping} and \textit{Dynamic Sampling}, which effectively mitigate entropy collapse and stabilize policy updates in long-CoT scenarios. 
The algorithm details and hyperparameter settings are provided in Appendix \cref{sec:grpo}.

\begin{table*}[th]
  \caption{Evaluation of reasoning capabilities and tool-integration efficiency on GSM8K test benchmark.
  We report the average@8 accuracy and tool-use frequency across simple and complex samples of each LLM. 
  $\spadesuit$ denotes frontier models accessing by API, $\heartsuit$ denotes RLVR-Trained models in TIR, $\clubsuit$ denotes OSS foundation models.
  \textit{Tool Freq.} denotes frequency of tool use for each example.
  In simple tasks, tool use slightly degrades performance, whereas in complex tasks, it substantially boosts accuracy.
  This finding highlights that tools are crucial for solving complex problems but should be avoided in simple ones.
  }
  \label{tab:pre1}
  \centering
  \small
  \begin{sc}
    \begin{tabular}{l|cccccc}
        \toprule
        \multirow{2}{*}{Model} & \multicolumn{3}{c}{Simple Samples (\%)} & \multicolumn{3}{c}{Complex Samples (\%)} \\
        \cmidrule(lr){2-4} \cmidrule(lr){5-7}
              & avg@8 & avg@8 w/ tool & Tool Freq. & avg@8 & avg@8 w/ tool & Tool Freq. \\
        \midrule
        $\spadesuit$ GPT-5.2       & 99.28 & 95.01\down{4.27} & 0.10 & 11.22 & 17.03\up{5.81} & 0.20\up{0.10} \\
        $\spadesuit$ Gemini-3      & 97.97 & 86.48\down{11.49} & 0.86 & 11.91 & 62.86\up{50.95} & 0.88\up{0.02} \\
        $\spadesuit$ Claude-4.5    & 70.98 & 72.84\up{1.86} & 0.44 & 23.38 & 68.54\up{45.16} & 0.33\down{0.11} \\
        $\spadesuit$ Deepseek-v3.2 & 99.29 & 94.18\down{5.11} & 0.98 & 9.01 & 20.06\up{11.05} & 1.26\up{0.28} \\
        $\spadesuit$ Deepseek-R1   & 96.44 & 79.66\down{16.78} & 0.68 & 16.92 & 52.84\up{35.92} & 0.69\up{0.01} \\
        $\spadesuit$ Glm-4.6       & 95.22 & 90.24\down{4.98} & 0.09 & 26.10 & 54.50\up{28.40} & 0.10\up{0.01} \\
        \rowcolor{blue!10} $\spadesuit$ Average  & 93.20 & 86.40\down{6.80} & 0.53 & 16.42 & 45.97\up{29.55} & 0.58\up{0.05} \\
        \midrule
        $\heartsuit$ SimpleTIR-7B  & 89.86 & 96.93\up{7.07} & 1.20 & 19.26 & 52.74\up{33.48} & 1.83\up{0.63} \\
        $\heartsuit$ ZeroTIR-7B    & 92.62 & 83.75\down{8.87} & 0.21 & 18.07 & 23.36\up{5.29} & 0.29\up{0.08} \\
        $\heartsuit$ ReTool-7B     & 92.27 & 87.21\down{5.02} & 1.19 & 23.04 & 28.02\up{4.98} & 1.25\up{0.06} \\
        $\heartsuit$ ReTool-32B    & 98.08 & 91.79\down{6.29} & 3.09 & 12.99 & 12.24\down{0.75} & 2.90\down{0.19} \\
        \rowcolor{blue!10} $\heartsuit$ Average  & 93.21 & 89.92\down{3.29} & 1.42 & 18.34 & 29.09\up{10.75} & 1.57\up{0.15} \\
        \midrule
        $\clubsuit$ Qwen3-8B         & 99.22 & 97.60\down{1.62} & 0.56 & 8.06 & 10.28\up{2.22} & 0.92\up{0.36} \\
        $\clubsuit$ Qwen3-32B        & 99.21 & 97.60\down{1.61} & 0.37 & 10.42 & 16.67\up{6.25} & 0.55\up{0.18} \\ 
        $\clubsuit$ Qwen2.5-7B-inst  & 96.99 & 91.66\down{5.33} & 1.03 & 13.38 & 27.50\up{14.12} & 0.98\down{0.05} \\
        $\clubsuit$ Qwen2.5-32B-inst & 98.52 & 97.07\down{1.45} & 1.47 & 7.50 & 12.50\up{5.00} & 1.58\up{0.11} \\
        $\clubsuit$ Llama3.1-7B-inst & 91.95 & 64.24\down{27.71} & 1.00 & 21.87 & 17.63\down{4.24} & 0.99\down{0.01} \\ 
        $\clubsuit$ Llama3.2-3B-inst & 88.58 & 39.43\down{49.15} & 0.70 & 23.23 & 8.87\down{14.36} & 0.58\down{0.12} \\ 
        \rowcolor{blue!10} $\clubsuit$ Average  & 95.75 & 81.27\down{14.48} & 0.86 & 14.08 & 15.58\up{1.50} & 0.93\up{0.07} \\
        \bottomrule
    \end{tabular}
  \end{sc}
\end{table*}

\section{Experiments: Quantifying Tool Overuse}
\label{sec:quantify}
In this section, we evaluate a diverse range of LLMs to quantify the \textit{tool overuse} problem in TIR tasks. 
To systematically analyze this phenomenon, we implement the following experimental protocol for each benchmark:
\begin{enumerate}
    \item {Base Reasoning}: The model solves problems through reasoning, and provides the final answer without relying on external tools. 
    \item {Task Categorization}: Using the $avg@8$ accuracy, we partition each test set into two subsets: \textit{Simple} (where $avg@8 \geq 0.5$) and \textit{Complex} (where $avg@8 < 0.5$). 
    This enables us to observe whether the use of the tools is related to the difficulty level of the problems.
    \item {Autonomous Tool Use}: The model is provided with tool-use capabilities and independently decides whether to invoke them.
    We report the $avg@8$ scores with tools to mitigate variances, and recorded the average number of tool invocations required for each question.

\end{enumerate}

\textbf{Models \& Benchmarks.} The evaluation covers models across three categories:
(1) \textit{Open-Source Models}: Qwen2.5 (7B/32B)-Instruct, Qwen3-8B, Llama3.1-8B, and Llama3.2-3B.
(2) \textit{RLVR-Trained Models}: We include Retool-7B/32B \citep{feng2025retool}, SimpleTIR \citep{xue2025simpletir} and ZeroTIR-7B \citep{mai2025agent}.
Notably, SimpleTIR particularly employs a feedback shielding mechanism to indirectly prevent the occurrence of ineffective tool use.
(3) \textit{Frontier Models}: Top-tier models including DeepSeek-V3.2, GPT-5.2, GLM-4.6, Gemini-3-Pro-Preview, and Claude-Haiku-4.5.
We evaluate on three math benchmarks: {GSM8K} for basic multi-step arithmetic, 
and {AIME24/25} for advanced Olympic-level mathematics. Details are in \cref{sec:benchmark}.

\subsection{Result on Redundant Tool Use}
Evaluation results on the GSM8K test benchmark are presented in \cref{tab:pre1}.
Firstly, we find that the frequency of tool use is consistently non-zero (0.93 times on average) on simple samples across all models, indicating that tool overuse is a widespread phenomenon.
Furthermore, the marginal difference in tool-use frequency between simple and complex samples ($\Delta \in [0.05, 0.15]$).
It suggests that current LLMs fail to strategically decide when a tool is actually necessary.

\begin{figure}[t]
  \vskip 0.1in
  \begin{center}
    \centerline{\includegraphics[width=0.85\columnwidth]{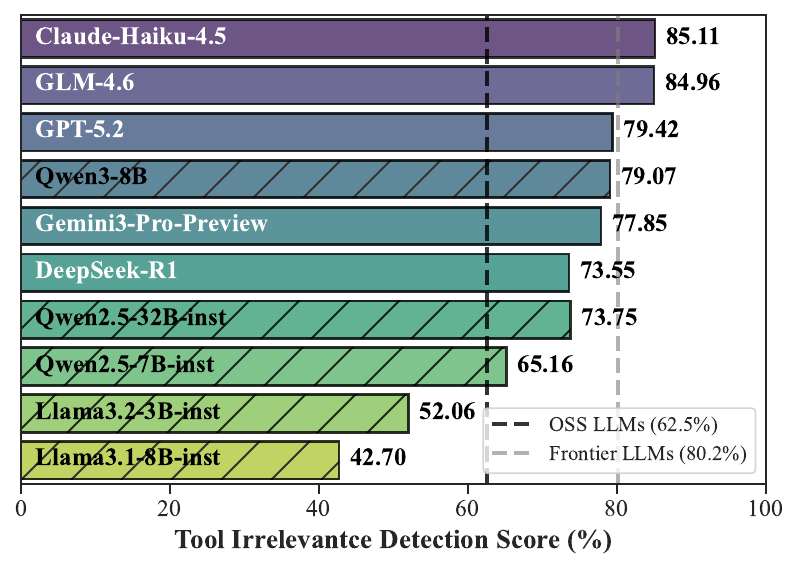}}
    \caption{
       The result of quantifying overuse in irrelevant tools. 
       Higher scores indicate greater precision in tool selection, with fewer irrelevant calls.
       Top-tier models achieve only 80.2\% performance on average, while open-source models get just 62.5\% on average.
       It highlights that irrelevant tool overuse remains a pervasive issue, even among frontier models.
    }
    \label{fig:irrel}
  \end{center}
  \vskip -0.3in
\end{figure}

Secondly, we surprisingly observe that tool-integrated reasoning harms performance on simple samples, those well within the model’s intrinsic capacity boundary. 
Concretely, on \textit{simple} samples, $avg@8$ with tool score drops by $3.29\%\sim14.49\%$ on average in various types of LLMs.
For instance, tool-use Gemini3’s $avg@8$ score decreases by 11.49\% with TIR compared to standard reasoning without tools. 
We attribute this issue to the fact that the invocation of those tools introduced unnecessary context overhead, without providing any additional useful information, thereby interfering with the natural reasoning process of the model.

Finally, tool overuse varies significantly across models. Among open-source foundations, the Llama series exhibits the most severe issues: 
tool invocation on simple samples incurs a 49.15\% performance penalty, while the gains on complex samples drop by 14.36\%, highlighting a lack of proficient tool-use logic. 
In contrast, the Qwen series perform more evenly. 
Notably, RLVR-trained models show significantly higher tool-use rates (e.g., increasing 65\% compared to OSS models), suggesting that RLVR may exacerbate tool overuse. 
Conversely, API-based frontier models invoke tools least frequently, reflecting superior baseline capabilities without external tools.

Additionally, statistics for \textit{simple} and \textit{complex} subsets in each model, along with their evaluations on AIME24 and AIME25 benchmark, are provided in Appendix ~\cref{sec:add_res_overuse}. 
These results corroborate the observations presented above.

\subsection{Results on Irrelevant Tool Use}
We identify another mode of overuse, \textit{irrelevant tool-use}.  
Specifically, we evaluate the cases where the provided tools are all irrelevant to the user's query, in which the model ought to refrain from invoking any tools whatsoever.  
Following BFCL~\citep{patil2025bfcl}, the standard benchmark for tool-use evaluation, we report irrelevance detection results.  
In this scenario, a correct response should either explain the tool's irrelevance or provide a standard text response without a function call.
As shown in ~\cref{fig:irrel}, top-tier models achieve only 80.2\% accuracy on average, indicating a 19.8\% tendency to spuriously invoke tools to compensate for internal reasoning gaps.  
This problem is more severe in open-source models, with error rates rising to 37.5\%.
These observations indicate that most models exhibit a tendency to employ irrelevant tools, a phenomenon that is likely caused by semantic illusions arising from insufficient reasoning capabilities, which in turn hinder problem-solving.

In summary, tool overuse is a widespread issue across models: 
it incurs unnecessary computational overhead, and even degrades performance on simple reasoning tasks. 
To facilitate future research in addressing this problem, the following section investigate the underlying causes of such overuse.

\begin{figure*}[ht]
  \vskip 0.2in
  \begin{center}
    \begin{subfigure}[t]{0.99\linewidth}
        \includegraphics[width=1.0\linewidth]{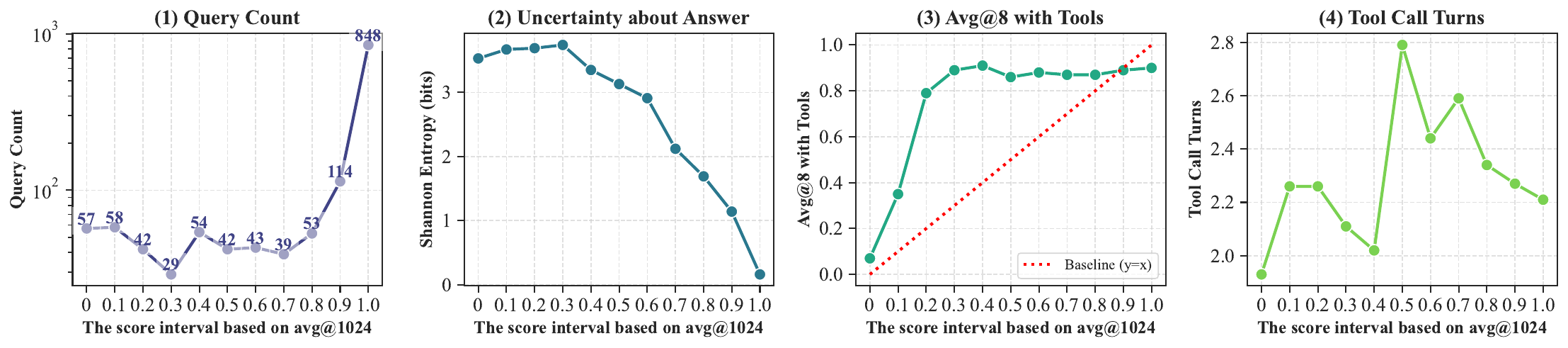}
        \caption{The evaluation results based on Qwen3-8B model.}
        \label{fig:qwen2.5}
    \end{subfigure}
  \hfill
    \begin{subfigure}[t]{0.99\linewidth}
        \includegraphics[width=1.0\linewidth]{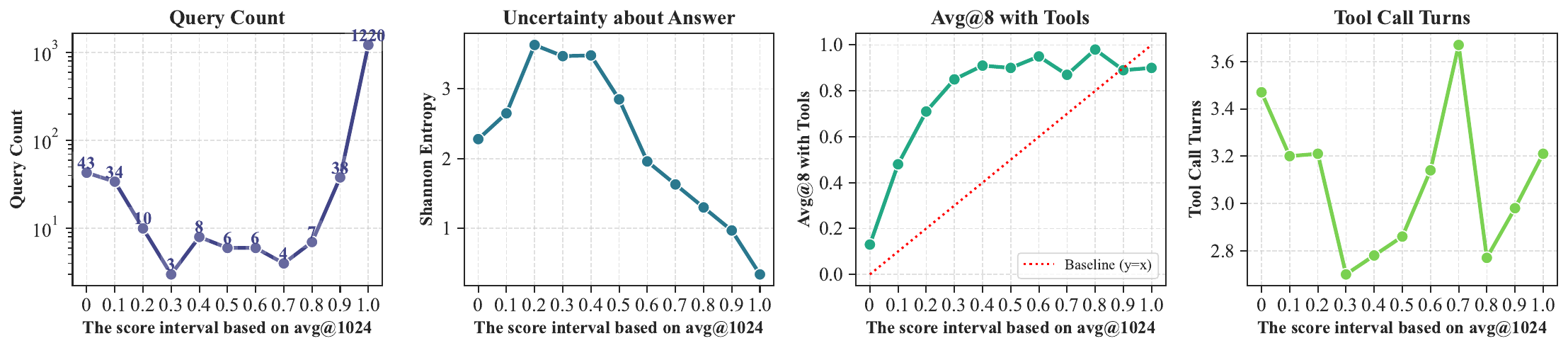}
        \caption{The evaluation results based on ReTool-32B model}
        \label{fig:retool-32b}
    \end{subfigure}
    \begin{subfigure}[t]{0.99\linewidth}
        \includegraphics[width=1.0\linewidth]{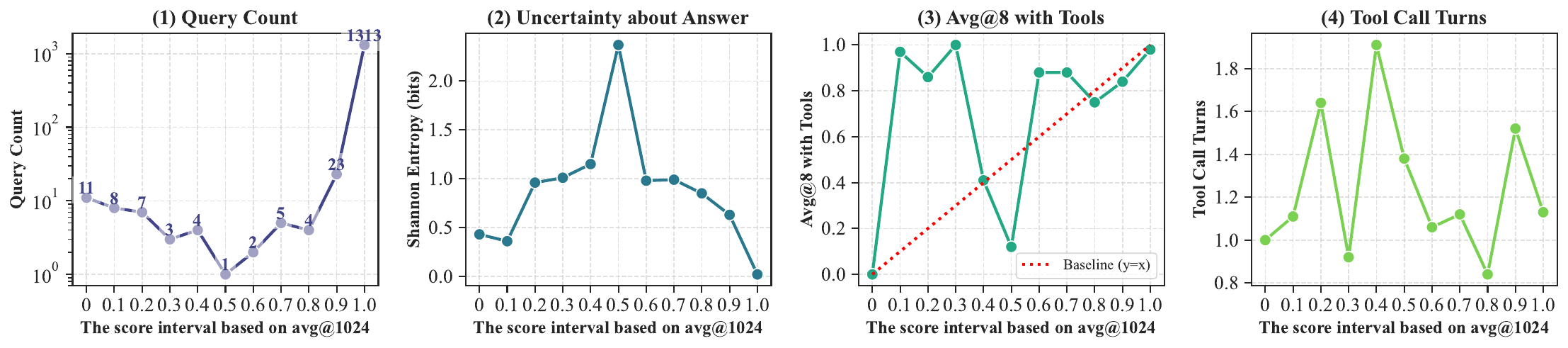}
        \caption{The evaluation results based on Gemini3.0-Pro-Preview model}
        \label{fig:gemini}
    \end{subfigure}
    \caption{
      Tool-use behavior and performance across the model’s internal knowledge availability.
    Contrary to the intuition that higher internal knowledge availability should correlate with fewer tool calls, the model exhibits a {knowledge epistemic illusion}. 
    Notably, performance degrades when tools are invoked in high-availability regions ($avg@1024 > 0.9$). 
    This suggests that models often misjudge their internal boundaries, leading to an over-reliance on external tools even when internal knowledge is sufficient.
    }
    \label{fig:boundary}
  \end{center}
  \vskip -0.1in
\end{figure*}

\section{Empirical Analysis A: Knowledge Epistemic Illusion}
\label{sec:passk}
We hypothesize that the \textit{knowledge epistemic illusion}, a phenomenon where models underestimate their internal knowledge as they approach their capacity limits, is a primary driver of tool overuse.
To verify this, we first measure each model’s internal knowledge availability.
This allows us to map and analyze how tool-use behaviors shift across different availability regions.

\subsection{Measuring Knowledge Availability}
\label{sec:boundary}
We compute the ${avg@k}$ metric (with a large $k$, e.g., 1024) to estimate the model’s intrinsic knowledge availability, inspired by \citet{yue2025does,sun2025rl}. 
\citet{yue2025does} show that base models achieve maximal reasoning coverage under high sampling budgets, making 
${avg@1024}$ a reliable proxy for knowledge availability.
Consequently, ${avg@1024}$ is positively correlated with the model’s intrinsic knowledge availability.
If ${avg@1024 > 0}$, the instance lies within the model’s ability boundary; otherwise, it falls outside (i.e., no knowledge availability). 
Scores equal to 0 represent beyond boundary cases, while values near 1 signify strong internal support (i.e., high knowledge availability).
Thus, ${avg@1024}$ serves as a quantitative measure of actual internal knowledge availability, enabling rigorous evaluation of the model’s reasoning capacity coverage.

We evaluate on a mixed test set comprising GSM8K, AIME 2024, and AIME 2025. 
For each input $x$ (without external tools), we generate $N = 1024$ independent answers $\{y^{(i)}\}_{i=1}^{N}$ via stochastic decoding (temperature = 1.0). 
Separately, we run 8 tool-integrated reasoning trials per problem. 
Based on the tool-free \texttt{avg@1024} score, we partition the queries into 11 bins. 
For each bin, we report: (1) the number of queries, 
(2) epistemic uncertainty about the generated answer.
(3) tool-augmented \texttt{avg@8} scores, 
and (4) tool-call turns per query (knowledge-behavior correlation).

The epistemic uncertainty in tool-free setting is quantified by the Shannon entropy ~\citep{gal2016dropout}, and computed as follows. 
Let $\hat{p}(y) = \frac{1}{N} \sum_{i=1}^{N} \mathbb{I}[y^{(i)} = y]$ be the empirical probability of answer $y$, where $\mathcal{Y}$ is the set of distinct generations. 
we calculate the Shannon entropy of the empirical distribution over unique answers:
\begin{equation}
H(Y \mid x) = -\sum_{y \in \mathcal{Y}} \hat{p}(y) \log_2 \hat{p}(y),
\end{equation}
where higher entropy indicates greater disagreement among generations, reflecting higher epistemic uncertainty. 

\subsection{Results Analysis}
As shown in \cref{fig:boundary}, we evaluate the performance of Gemini, Qwen3-8B, and ReTool-32B.
Results for other models are provided in the appendix~\cref{sec:add_res_boundary}.
Our analysis yields the following key insights:

First, \textbf{correlation between performance and uncertainty}. 
For more than half of the queries, the models achieve an $avg@1024>0.9$.
Globally, we observe that answer uncertainty decreases as $avg@1024$ increases.
Locally, however, the greatest benefit from tool use occurs in high-uncertainty regions: 
for Qwen3-8B and ReTool-32B, this lies in $avg@1024\in(0.2,0.4)$, and for Gemini-3, near $(0.1,0.3)$, 
both showing significantly higher gains over the no-tool baseline ($y=x$).
This indicates that tool use effectively mitigates uncertainty in tool-free settings, thereby boosting overall performance.

However, \textbf{the integration of external tools does not always lead to better outcomes}.
Interestingly, tool integration does not yield uniform benefits. 
When a query falls well within the model’s internal knowledge boundary, invoking tools often degrades performance. 
For example, in Qwen3-8B, this critical threshold occurs at $avg@1024 \approx 0.8$.
This suggests overusing tools on simple tasks can be counterproductive.

Furthermore, \textbf{the models favor tools even when they possess sufficient internal knowledge}.
In cases where $avg@1024 > 0.9$, the tool-calling rate remains unexpectedly high, surpassing even that in low-knowledge regions.
Qwen3-8B, for example, averages 2.2 tool calls per query in this range. 
Conversely, when internal knowledge is virtually absent (\( \text{avg@1024} < 0.2 \)), some model (e.g., Qwen3-8B and Gemini3) often fails to invoke tools at all. 
We attribute this to severe hallucination, where the model erroneously believes it has already solved the problem correctly.
These observation shows models rely on external tools by default, often overlooking perfectly accurate internal information.


Most surprisingly, LLMs exhibit a counterintuitive pattern: \textbf{higher internal knowledge availability does not lead to fewer tool calls.}
We do not observe a consistent correlation between tool-use behavior and internal knowledge availability. 
On a global scale, the frequency of tool invocations remains relatively stationary even as $avg@1024$ increases.
This suggests that models fail to align their perceived knowledge boundary with their actual capability, instead triggering tools excessively.
We term this phenomenon knowledge epistemic illusion: a hallucinatory misperception of internal knowledge availability.
It reflects a lack of awareness to assess and utilize internal knowledge, leading to blind reliance on external tools.



\subsection{Mitigation Insight}
The knowledge epistemic illusion suggests a key implication: aligning the model’s perceived knowledge with its actual knowledge capacity could reduce unnecessary tool reliance.
To validate this hypothesis, we propose a knowledge-aware alignment strategy based on direct preference optimization (DPO \citep{rafailov2024direct}),
which explicitly encourages the model to trust its internal knowledge when appropriate, and calls tools only near genuine knowledge boundary.
Specifically, we perform 1,024 independent reasoning trials on the GSM8K training set. 
Among these, we curated preference pairs by contrasting trajectories with minimal tool calls (e.g., 0 or 1 tool call) against those with excessive usage under identical prompts.
Then, DPO update the policy model by pulling the policy toward preferred responses and away from dispreferred ones; 
details are provided in Appendix \cref{sec:dpo}.

\begin{figure}[th]
  \begin{center}
    \begin{subfigure}[t]{1.0\linewidth}
        \includegraphics[width=1.0\linewidth]{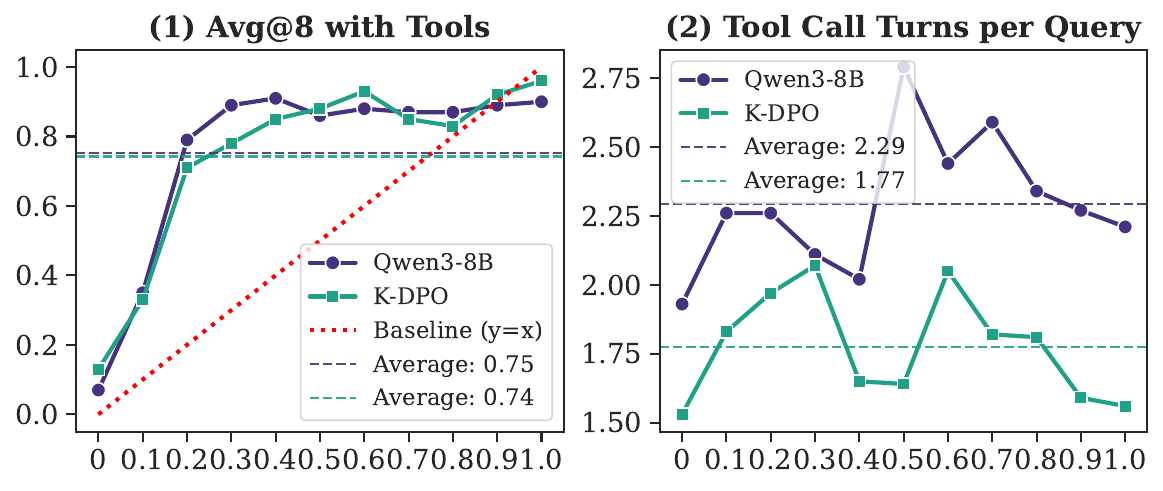}
        \caption{The evaluation results based on Qwen3-8B model.}
        \label{fig:dpo11}
    \end{subfigure}
        \begin{subfigure}[t]{1.0\linewidth}
        \includegraphics[width=1.0\linewidth]{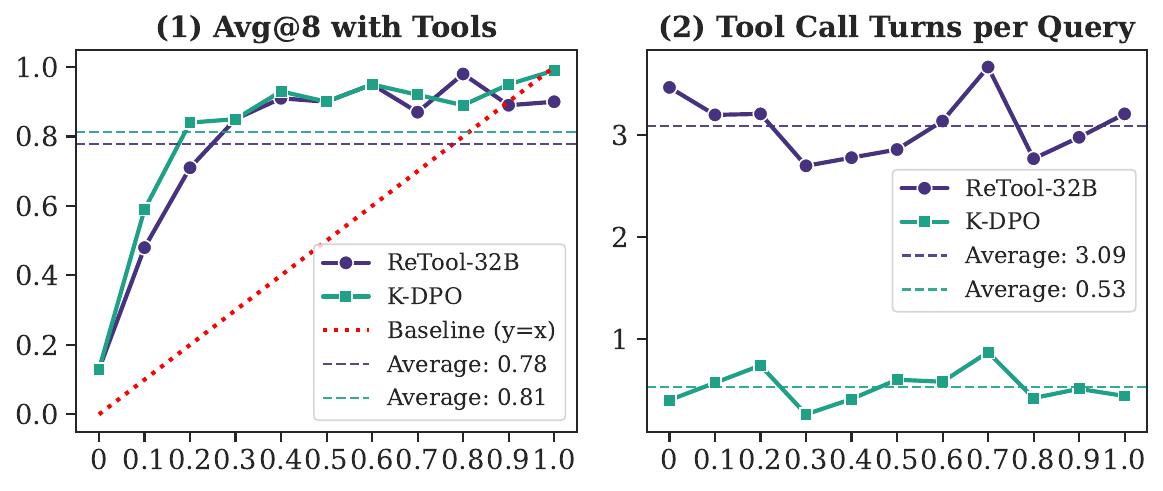}
        \caption{The evaluation results based on ReTool-32B model.}
        \label{fig:dpo22}
    \end{subfigure}
    \caption{
    Evaluation results comparing the base model and our K-DPO trained model. 
    We report (1) the Avg@8 score with tools and (2) the knowledge–behavior correlation curve (measured by tool-call turns). 
    Our approach reduces tool-call turns in higher $avg@1024$ ranges while improving overall tool-augmented performance.
    }
    \label{fig:dpo_res}
  \end{center}
  \vskip -0.3in
\end{figure}

As illustrated in \cref{fig:dpo_res}, our Knowledge-aware DPO (K-DPO) strategy consistently improves performance when tools are integrated. 
Specifically, in (1) avg@8 with tools, we observe an sightly increase in average $avg@8$ scores by 1\% for Qwen3-8B and 3\% for ReTool-32B.
Notably, our strategy elevates the benefit threshold for tool usage (e.g., from 0.8 to 0.95 in Qwen3-8B; from 0.9 to 1.0 in ReTool-32B).
Moreover, the knowledge–behavior correlation curve (2) further reveals that the above improvement is associated with a reduction in tool-call turns.
For instance, the average number of tool-call turns decreases by 0.52 ($\downarrow$22.7\%) in Qwen3 and 2.56 ($\downarrow$82.8\%) in ReTool-32B.
These results suggest that, reducing unnecessary tool-turns in high internal knowledge availability cases can effectively enhance performance. 
This is may because fewer tool-call turns reduce the contextual length burden without sacrificing any informational gain.
Additionally, evaluations across three separate benchmarks (~\cref{tab:overall}) demonstrate that our method improves accuracy while reducing tool-use turns.
These finding proves that aligning the model’s perceived knowledge with its actual one can reduce tool-overuse.


\section{Experiment B: Outcome-Reward Trap}
\label{sec:reward}
Given that RLVR is the primary method for tool-integration training, we focus on a specific issue: RLVR-trained models exhibit an unusually high tool overuse rate compared to OSS models, as shown in~\cref{tab:pre1}.
We hypothesize that the outcome-only reward design in RLVR training is a key driver of this behavior. 
To test this, we first visualize the training process to demonstrate this causal link and then provide insights for potential mitigation strategies.

\begin{figure}[th]
  \begin{center}
    \centerline{\includegraphics[width=0.7\columnwidth]{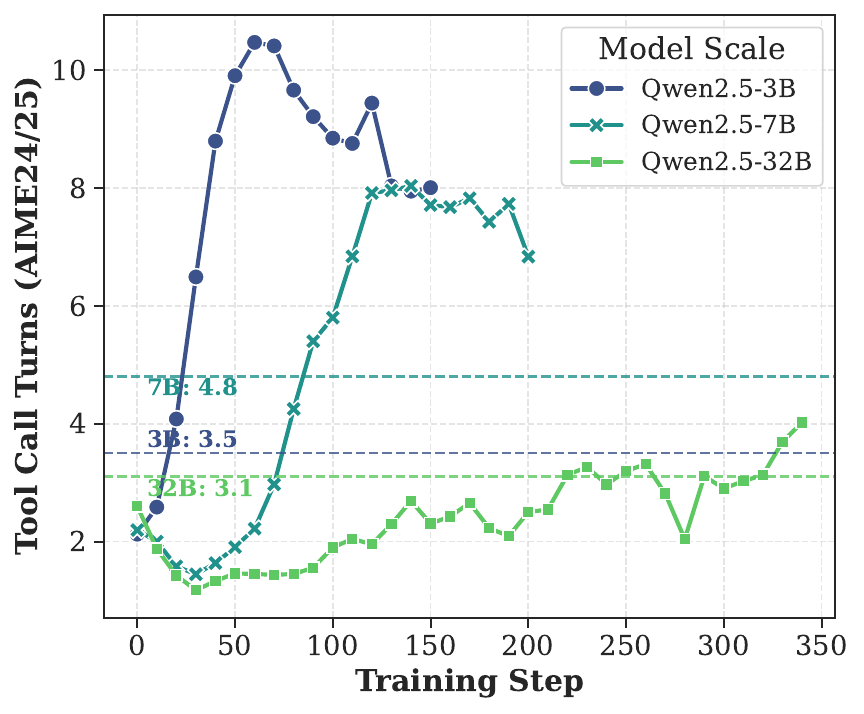}}
    \caption{
    Training dynamics of tool-call turns across model sizes.
    As RLVR training progresses, models increasingly overuse tools. 
    The dashed line denotes the reasonable upper bound (tool@1024), which is consistently exceeded.
    }
    \label{fig:training}
  \end{center}
  \vskip -0.3in
\end{figure}

\subsection{Visualizing the Training Process}
\label{sec:training}
We visualize the training of Qwen2.5-3B/7B/32B-Instruct following ReTool~\citep{yu2025dapo}, i.e., SFT first then RLVR. 
Specifically, we report the number of tool-call turns on the validation set (AIME24/25), trained on DAPO-Math-17k with RLVR; 
The maximum allowed tool-call turns is 16.
To estimate the upper bound of tool-use frequency, we introduce ${tool@1024}$: the average maximum number of tool calls required to reach a correct answer per sample. 
We compute the tool@1024 score on the validation set for each model fine-tuned from the base model via SFT.
It denotes the theoretical upper bound of rational tool usage, which a model can achieve while successfully resolving a task.

As shown in \cref{fig:training}, we observe that:
(1) The number of tool calls increases significantly during RLVR training, for example, from 2.2 to 6.8 turns in the Qwen2.5-7B model.
(2) The outcome-reward trap is more pronounced in smaller models (e.g., 3B), which rapidly increase tool usage, suggesting limited tool-use efficiency.
(3) Compared to the tool@1024 metric, trained models consistently exceed the reasonable upper bound on tool calls, indicating that RLVR shifts the model’s empirical distribution and substantially boosts tool-use frequency.
Collectively, these observations show optimizing solely for final correctness, while ignoring procedural inefficiency, encourages excessive tool use.

\subsection{Ablation Verification}
\label{sec:outcome}
To further verify the outcome-reward trap, we conduct an ablation using an outcome-efficiency tradeoff reward in inter-group sampling of DAPO.
Specifically, we increase the rollout group size to 32 to better approximate the ideal policy, i.e., correct answers with minimal tool usage. 
We treat this ideal as the ground truth (assigning a +1 reward). 
For responses with the correct final answer, we apply an additional penalty of 0.05 per tool call (capped at 16 turns) alongside the standard outcome reward (±1).
Under this setting, we train our model starting directly from RLVR-trained models.
Experiment details see appendix \cref{sec:exp_details}.
                         
\begin{figure}[th]
  \begin{center}
    \centerline{\includegraphics[width=0.7\columnwidth]{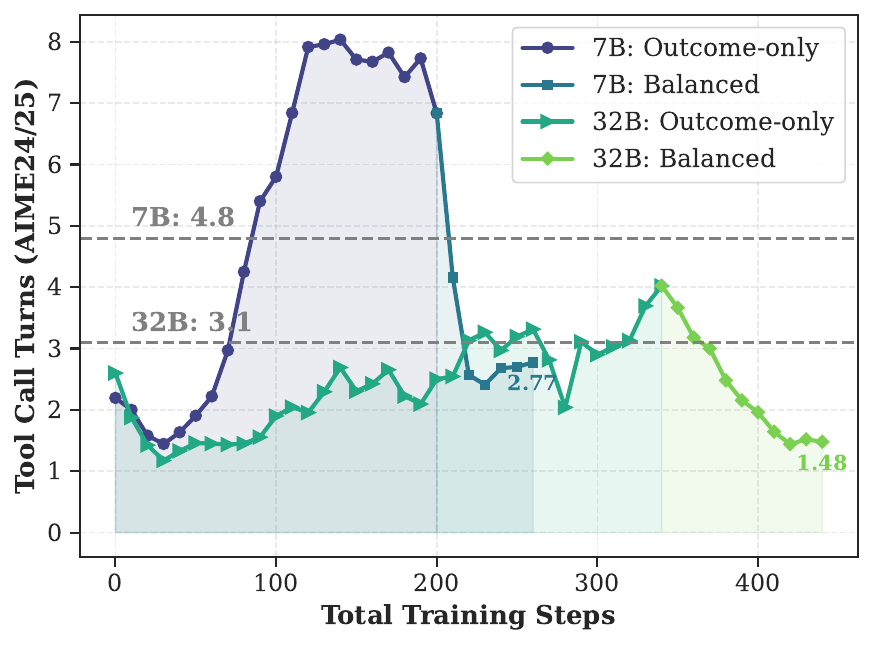}}
    \caption{
    Training dynamics of tool-call turns under two reward schemes in 7B/32B-scale model.
    As RLVR training progresses, our balanced reward decrease the tool-call turns compared to the outcome-only reward.
    }
    \label{fig:qwen7b}
  \end{center}
  \vskip -0.3in
\end{figure}

The training visualization results for the 7B-scale and 32B-scale models are reported in \cref{fig:qwen7b}.
Their evaluation score dynamics are presented in Appendix \cref{sec:add_res_training}.
Under continued RLVR training, our balanced outcome-efficiency reward significantly reduces tool-call turns compared to the outcome-only reward.
For example, our outcome-efficiency reward in 7B-scale reduces tool-call turns by 59.4\% (from 6.83 to 2.77) on average.

We further evaluated model performance with tools across three benchmarks, in different training methods, i.e., 
$\diamondsuit$our knowledge-aware DPO(\textbf{K-DPO}), $\diamondsuit$RLVR-trained models (\textbf{outcome}-only reward and our \textbf{balanced} outcome-efficiency reward).
As summarized in \cref{tab:overall},
Notably, our balanced outcome-efficiency reward achieves better performance comparable to outcome-only reward. 
For instance, it yields an average improvement of -1.1\% in 7B-scale model and +0.9\% in 32B-scale model.
Moreover, it also reducing tool usage during evaluation, e.g., 66.7\% (from 5.1 to 1.7) turns in 7B-scale model and 60.7\% (from 2.8 to 1.1) turns in 32B-scale model.
These results demonstrate that the balanced reward significantly reduces tool-call frequency during problem-solving without compromising evaluation accuracy.
Additional ablation studies regarding rollout group size and sampling temperature are provided in \cref{sec:ablation}.

\begin{table}[th]
  \caption{Performance comparison of tool-augmented models across three benchmarks. 
  \textbf{Bold} indicates the highest accuracy (avg@8); subscripts denote tool-call turns, and its \underline{underlining} marks the lowest value.
  }
  \label{tab:overall}
  \small
  \begin{center}
    \begin{small}
      \begin{sc}
        \begin{tabular}{l|ccc}
          \toprule
            {Model} & {GSM8K}\textsubscript{test} & {AIME24} & {AIME25} \\
            \midrule
            {Qwen2.5-7B} 
                 & 60.71\textsubscript{1.5}  & 10.42\textsubscript{\underline{1.4}} & 2.92\textsubscript{\underline{1.5}} \\
             \hspace{0.2cm} + $\heartsuit$Outcome & \textbf{92.34}\textsubscript{2.2}  & 40.00\textsubscript{4.5 } & 23.33\textsubscript{8.6} \\
             \hspace{0.2cm}  + $\heartsuit$Balanced & 92.44\textsubscript{\underline{0.3}} & \textbf{35.83}\textsubscript{2.3} & \textbf{24.17}\textsubscript{2.5} \\
            \midrule
            {Qwen2.5-32B} 
                 & 91.39\textsubscript{1.0}  & 23.33\textsubscript{1.4} & 15.0\textsubscript{\underline{1.4}} \\
             \hspace{0.2cm} + $\heartsuit$Outcome & 94.14\textsubscript{0.5}  & 56.89\textsubscript{3.2} & \textbf{54.67}\textsubscript{4.1} \\
             \hspace{0.2cm} + $\heartsuit$Balanced & \textbf{94.90}\textsubscript{\underline{0.1}}  & {59.17}\textsubscript{1.5} & \textbf{53.33}\textsubscript{{1.7}} \\
             
            \midrule
            {Qwen3-8B} 
                & 86.13\textsubscript{2.2}  & 43.33\textsubscript{1.9} & 26.7\textsubscript{\underline{1.2}} \\
                 \hspace{0.2cm} + $\diamondsuit$K-DPO & {89.61}\textsubscript{{1.6}}  & {33.33}\textsubscript{{1.4}} & {33.33}\textsubscript{1.5} \\
            {Retool-32B} 
                & 87.26\textsubscript{2.5}  & 50.0\textsubscript{3.1} & 36.67\textsubscript{3.1} \\
                 \hspace{0.2cm} + $\diamondsuit$K-DPO & \textbf{94.16}\textsubscript{\underline{0.5}}  & \textbf{53.33}\textsubscript{\underline{1.2}} & \textbf{43.33}\textsubscript{2.5} \\
         \bottomrule
        \end{tabular}
      \end{sc}
    \end{small}
  \end{center}
  \vskip -0.1in
\end{table}

\section{Theoretical Analysis: The Utility Landscape of Tool Use}
\label{sec:theoretical}

\subsection{Theoretical Framework}
\label{sec:framework}
\paragraph{Problem Formulation.}
We formalize the inference process of tool-augmented LLMs as a binary decision-making problem.
Given an input query $x$, the model policy $\pi_\theta$ selects an action from the space $\mathcal{A} = \{\mem, \tool\}$:
\begin{itemize}
    \item $\mem$ ({Internal Retrieval}): Generates response $y$ relying solely on parametric memory.
    \item $\tool$ ({External Invocation}): Invokes an external tool to acquire intermediate results before generating $y$.
\end{itemize}
The training objective is to maximize the expected utility $U$:
\begin{equation}
    a^* = \operatorname*{arg\,max}_{a \in \mathcal{A}} \; \mathbb{E}_{y|x,a} \left[ U(y, a) \right]
\end{equation}

\begin{definition}[Tool-Use Utility]
\label{def:utility}
In standard settings, reward functions often prioritize correctness over efficiency. To analyze tool overuse, we decompose the utility function into \textit{Performance Gain} and \textit{Inference Cost}:
\begin{equation}
    U(y, a) = \mathcal{R} \cdot \mathbbm{1}(y = y^*) - \lambda \cdot \mathcal{C}(a)
\end{equation}
where $\mathcal{R} > 0$ is the scalar reward for a correct response, $\mathbbm{1}(\cdot)$ is the indicator function for the ground truth $y^*$, $\mathcal{C}(a)$ represents the cost (e.g., latency, token count or number of tool calls), and $\lambda$ is the coefficient of efficiency sensitivity.

\end{definition}

\paragraph{The Decision Boundary.}
The model selects $\tool$ if and only if the expected utility of tool use exceeds that of internal retrieval: $\mathbb{E}[U(\tool)] > \mathbb{E}[U(\mem)]$.
Let $P(y^* | x, a)$ denote the ground-truth probability of correct response given action $a$. 
The invocation condition is derived as:
\begin{align}
    P(y^* | x, \tool) \cdot \mathcal{R} - \lambda \mathcal{C}_{\text{tool}} &> P(y^* | x, \mem) \cdot \mathcal{R} - \lambda \mathcal{C}_{\text{mem}} \notag
\end{align}
\begin{proposition}[Tool Invocation Condition]
\label{prop:boundary}
The optimal policy invokes a tool if and only if the Marginal Reliability Gain exceeds the Marginal Cost:
\begin{equation}
    \underbrace{P(y^* | x, \tool) - P(y^* | x, \mem)}_{\Delta P \text{ (Marginal Reliability Gain)}} > \frac{\lambda}{\mathcal{R}} \cdot \underbrace{(\mathcal{C}_{\text{tool}} - \mathcal{C}_{\text{mem}})}_{\Delta C \text{ (Marginal Cost)}}
    \label{eq:boundary}
\end{equation}
\end{proposition}
\cref{eq:boundary} establishes the theoretical lower bound for justifiable tool-use.

\subsection{Theoretical Causes of Overuse}
By identifying pathological conditions on \cref{prop:boundary}, we demonstrate why current LLMs inevitably converge towards tool overuse. 

\paragraph{Proposition 1: The Structural Utility Imbalance (Rational Overuse).}
\textit{Derivation:} In existing Supervised Fine-Tuning (SFT) and RLVR paradigms, models are optimized solely for final response accuracy without penalty for process inference cost. 
This implies the efficiency coefficient $\lambda \to 0$, the reward scale such that $\mathcal{R} \to 1$ (standard binary classification utility).
Consequently, the cost term on \cref{eq:boundary} vanishes:
\begin{equation}
    \lim_{\substack{\mathcal{R} \to 1 \\ \lambda \to 0}} \frac{\lambda}{\mathcal{R}} \cdot \Delta C = 0 \quad \implies \quad \Delta P > 0
\end{equation}
\textit{Implication:} As long as the tool offers a non-zero probability improvement (e.g., $\Delta P = 10^{-4}$), tool invocation becomes the global optimum. 
For a LLM incurring no penalty for compute, overuse is rational.

\paragraph{Proposition 2: Miscalibrated Confidence (Epistemic Illusion).}
The model's action decision relies not on the true probability $P$, but on the model's estimate $\hat{P}_\theta$. 
Experiment \cref{sec:passk} shows that the model underestimates the availability of its internal knowledge and over-relies on tool use.
Therefore, we observe two distinct calibration failures:
\begin{enumerate}
    \item \textbf{Under-confidence in Memory Knowledge:} underestimates its internal knowledge, i.e., $\hat{P}_\theta(y^* | x, \mem) \ll P_{\text{true}}(y^* | x, \mem)$.
    \item \textbf{Over-confidence in Tools:} blindly trusts external outputs, i.e., $\hat{P}_\theta(y^* | x, \tool) \approx 1$.
\end{enumerate}
This results in an inflated marginal gain: $\Delta \hat{P}_\theta \gg \Delta P_{\text{true}}$ in ~\cref{eq:boundary}.
\textbf{This indicates a significant calibration gap where the model's perceived knowledge is far lower than its actual knowledge.}
This calibration gap leads the epistemic illusion about internal knowledge availability.

\textbf{In summary}, tool overuse emerges from the interplay between epistemic illusion about intrinsic knowledge availability and utility imbalance in RLVR reward. 
The former causes models to underestimate internal knowledge availability, and the latter collapses the utility function by ignoring inference costs. 

\section{Conclusion}
In this work, we systematically investigate the tool overuse phenomenon, uncovering its root causes and offering practical implications for future research.
Our analysis reveals that LLMs suffer from a severe {knowledge epistemic illusion}, leading them to underestimate their internal knowledge availability.
Besides, the {outcome-only reward} in RLVR training reinforces this tool overuse problem.
This pioneering study on tool overuse redefines the understanding of LLM autonomy, highlighting that reliable AI agents must balance internal reasoning with external tool use.


\section*{Impact Statement}

This paper presents work whose goal is to advance the field of Machine Learning.
There are many potential societal consequences of our work, none
which we feel must be specifically highlighted here.

\nocite{langley00}

\bibliography{example_paper}
\bibliographystyle{icml2026}

\newpage
\appendix
\onecolumn

\section{LLM USAGE}
During the preparation of this work, we used LLM to aid in polishing the manuscript and improving language. 
All final content was reviewed and revised by the authors to ensure its accuracy and originality. 
The core ideas, methods, and conclusions of the paper are solely the work of the authors.

\section{Limitations and Discussion}
First, our experiments focus exclusively on code-as-tool scenarios (Python interpreters) for mathematical reasoning. While this setting enables precise measurement of computational redundancy, our findings may not fully generalize to other tool types (e.g., search engines, APIs).

Second, we estimate internal knowledge availability via the ${avg@k}$ metric ($k=1024$), following \citet{yue2025does,sun2025rl}. 
This metric reflects a model's intrinsic capability ceiling under stochastic decoding and is justified by evidence that RL training rarely expands novel reasoning capacity beyond the base model's coverage \citep{yue2025does}. 
However, $avg@1024$ cannot pinpoint an exact knowledge-threshold for optimal tool invocation. 
As a practical proxy, we adopt the monotonicity principle: \textbf{higher internal knowledge availability should correlate with fewer tool calls.}
Future work could explore dynamic thresholding mechanisms that adapt to task-specific uncertainty.


\section{Detailed RL Algorithms}

\subsection{RLVR Algorithms: GRPO and DAPO}
\label{sec:grpo}
Group Relative Policy Optimization~\citep{shao2024deepseekmath} efficient optimize the model's reasoning capabilities without the significant memory overhead of a Value Network.
Unlike Proximal Policy Optimization~\cite{schulman2017prox}, which relies on a separate value function to estimate the baseline, GRPO leverages the distribution of rewards within a group of sampled outputs to compute the advantage.

\paragraph{Group Sampling and Advantage Estimation.}
For each query $q$, GRPO samples a group of $G$ outputs $\{o_1, o_2, \dots, o_G\}$ from the current policy $\pi_\theta$. Let $r_i$ be the reward for the $i$-th output. The advantage $A_i$ is computed by normalizing the reward against the group statistics:
\begin{equation}
    A_i = \frac{r_i - \bar{r}}{\sigma_r + \epsilon}
\end{equation}
where $\bar{r}$ and $\sigma_r$ are the mean and standard deviation of the rewards within the group, respectively. This group-based baseline serves as a dynamic reference, reducing variance without requiring a learned value function.

\paragraph{Objective Function.}
The optimization objective maximizes the surrogate loss with a KL-divergence penalty to maintain stability relative to the reference model $\pi_{\text{ref}}$:
\begin{equation}
    \mathcal{J}(\theta) = \mathbb{E}_{q \sim \mathcal{D}, \{o_i\} \sim \pi_{\theta_{\text{old}}}} \left[ \frac{1}{G} \sum_{i=1}^G \left( \mathcal{L}_{\text{clip}}(r_t(\theta), A_i) - \beta \mathbb{D}_{\text{KL}}(\pi_\theta || \pi_{\text{ref}}) \right) \right]
\end{equation}
Here, $r_t(\theta) = \frac{\pi_\theta(o_i|q)}{\pi_{\theta_{\text{old}}}(o_i|q)}$ denotes the probability ratio, and $\mathcal{L}_{\text{clip}}$ is the standard PPO clipping operator:
\begin{equation}
    \mathcal{L}_{\text{clip}}(r_t, A_i) = \min \left( r_t A_i, \text{clip}(r_t, 1-\epsilon, 1+\epsilon) A_i \right)
\end{equation}
By eliminating the critic model, GRPO significantly reduces computational costs, enabling efficient training of large-scale reasoning models.

\textbf{Decoupled Clip and Dynamic sAmpling Policy Optimization (DAPO)}.
While GRPO offers efficiency, it often suffers from entropy collapse and training instability when dealing with complex reasoning and tool-call sequences. 
To address this, we employ DAPO as the RLVR algorthm in this paper.
It introduces four key techniques to make RL shine in the long-CoT RL scenario: 
(1) Clip-Higher, which promotes the diversity of the system and avoids entropy collapse;
(2) Dynamic Sampling, which improves training efficiency and stability;
(3) Token-Level Policy Gradient Loss, which is critical in long-CoT RL scenarios;
(4) Overlong Reward Shaping, which reduces reward noise and stabilizes training.

Totally, DAPO samples a group of outputs $\{o_i\}_{i=1}^G$ for each question $q$ paired with the answer $a$, and optimizes the policy via the following objective:

\begin{equation}
\begin{aligned}
\mathcal{J}_{\text{DAPO}}(\theta) = & \mathbb{E}_{(q,a) \sim \mathcal{D}, \{o_i\}_{i=1}^G \sim \pi_{\theta_{\text{old}}}(\cdot|q)} \\
&\left[ \frac{1}{\sum_{i=1}^G |o_i|} \sum_{i=1}^G \sum_{t=1}^{|o_i|} \min \left( r_{i,t}(\theta) \hat{A}_{i,t}, \, \text{clip}(r_{i,t}(\theta), 1 - \varepsilon_{\text{low}}, 1 + \varepsilon_{\text{high}}) \hat{A}_{i,t} \right) \right] \\
\text{s.t.} & \quad 0 < \left| \{o_i \mid \text{is\_equivalent}(a, o_i)\} \right| < G,
\end{aligned}
\end{equation}
where
\begin{equation}
r_{i,t}(\theta) = \frac{\pi_\theta(o_{i,t} \mid q, o_{i,<t})}{\pi_{\theta_{\text{old}}}(o_{i,t} \mid q, o_{i,<t})}, \quad \hat{A}_{i,t} = \frac{R_i - \text{mean}(\{R_i\}_{i=1}^G)}{\text{std}(\{R_i\}_{i=1}^G)}.
\end{equation}

\subsection{Direct Preference Optimization}
\label{sec:dpo}
Instead of learning an explicit reward model \citep{ouyang2022training}, Direct Preference Optimization (DPO) \citep{rafailov2024direct} reparameterizes the reward function \( r \) using a closed-form expression with the optimal policy:
\begin{equation}
r(x,y) = \beta \log \frac{\pi_{\theta}(y \mid x)}{\pi_{\text{ref}}(y \mid x)} + \beta \log Z(x),
\end{equation}
where \( \pi_{\theta} \) is the policy model, \( \pi_{\text{ref}} \) is the reference policy, \( Z(x) \) is the partition function, and the hyperparameter \(\beta\) scales the KL constraint. 
 Using the shorthand 
\(
h_{\pi_{\theta}}^{y_{w}} = \log \frac{\pi_{\theta}(y_{w} \mid x)}{\pi_{\text{ref}}(y_{w} \mid x)},  h_{\pi_{\theta}}^{y_{l}} = \log \frac{\pi_{\theta}(y_{l} \mid x)}{\pi_{\text{ref}}(y_{l} \mid x)}, 
\) 
at the \(i\)-th iteration, given a batch of preference data \(\mathcal{D}_{i}\) sampled with the latest policy \(\pi_{\theta(i-1)}\), we denote the policy objective \(\ell_{i}(\theta)\) as follows:
\begin{equation}
\small
\ell_{i}(\theta) = -\mathbb{E}_{(x, y_{w}, y_{l}) \sim \mathcal{D}_{i}} \left[  \log \sigma \left(\beta h_{\pi_{\theta}}^{y_{w}} -\beta h_{\pi_{\theta}}^{y_{l}}\right) \right],
\end{equation}
where \(y_{w}\) and \(y_{l}\) represent the step-level preferred and dispreferred responses, respectively.

\section{Related Work}
\textbf{Tool-Integrated Reasoning.}
It enables large language models to leverage external tools such as code interpreters in order to overcome the limitations of pure internal reasoning.
Early methods relied on prompt engineering \citep{chenprogram,yang2024buffer} and supervised fine-tuning (SFT) \citep{gou2023tora,yang2024qwen2,wang2023mathcoder}.
However, SFT-driven approaches often merely imitate demonstrations, lacking the flexibility required for robust agentic behavior.
Recent work, termed \textit{Agentic RL}, leverages RLVR to train more capable models \citep{guo2025deepseek,yu2025dapo,yu2025demystifying,zeng2025tool}.
By explicitly modeling tool execution within the action space and optimizing policy model via outcome-driven rewards, these methods enable LLMs to transcend static supervision and achieve robust reasoning.
Methods such as ReTool \citep{feng2025retool}, ToRL \citep{li2025torl}, and SimpleTIR \citep{xue2025simpletir} have been introduced to enhance TIR.
While these methods successfully improve accuracy via outcome-driven rewards, the resulting tool-call inefficiency remains underexplored.


\textbf{Tool Overuse.} 
The integration of external tools enables LLMs to transcend their internal knowledge boundaries \citep{chen2024unlocking,chen20252}. 
However, this capability introduces the challenge of \textit{tool overuse}, the execution of tool calls that are either computationally redundant or semantically irrelevant. 
Existing literature primarily focuses on the \textit{tool selection} task, evaluating an LLM's ability to determine when and which tools are required to achieve a correct response \citep{ning2024wtu,huang2024metatool,li2023api}. 
Their critical distinction: while tool selection prioritizes \textit{accuracy}, tool overuse addresses \textit{necessity}. 
For instance, a model may exhibit "correct selection but overuse," where the appropriate tool is identified but employed unnecessarily. 
A few studies in prompt engineering and SFT have touched upon reasoning efficiency and tool-use costs \citep{wang2025self,qian2025smart}.
Under RLVR training paradigm, SimpleTIR \citep{xue2025simpletir} incorporates a feedback masking mechanism to filter out trajectories containing void turns during RLVR, which indirectly suppresses the invalid tool use.
However, a comprehensive experimental analysis of tool overuse remains absent. 
This work addresses this blind spot in tool-integrated reasoning, 
recognizing that even if the final response is correct, the tool invocation decisions can be structurally unnecessary.

\section{Implementation Details}
\label{sec:exp_details}

\subsection{Train Setting}
RLVR Implementation: We implemented DAPO using the \texttt{verl} framework. 
Code execution was hosted in a sandboxed Python environment for security and reproducibility. 
For 7B-scale models, each RL session was completed within 24 hours on a cluster of 64 Ascend 910B NPUs (8 nodes). 
We reproduced the ReTool models in our local environment and observed a slight performance degradation compared to the results reported in the original paper.
We selected the checkpoint at step 200 for our final evaluations (see Table~\ref{tab:dapo_config} for details).

DPO Training: To facilitate knowledge-aware DPO, we curated preference pairs by contrasting trajectories with minimal tool calls (e.g., 0 or 1 tool call) against those with excessive usage under identical prompts. 
Following the standard DPO objective \citep{rafailov2024direct}, training was conducted for 1 epochs using LLaMA-Factory~\citep{zheng2024llamafactory}, converging in approximately 6 hours. 
To ensure stability, we averaged the performance across checkpoints in steps 20, 40, and 60. 
DPO hyperparameters are listed in Table~\ref{tab:dpo_config}.

\begin{table*}[ht]
    \centering
    \small
    \begin{minipage}{0.48\textwidth}
        \centering
        \caption{Hyperparameters for DAPO training.}
        \label{tab:dapo_config}
        \begin{sc}
        \begin{tabular}{lc}
            \toprule
            \textbf{Hyperparameter} & \textbf{Value} \\
            \midrule
            \rowcolor{gray!10} \multicolumn{2}{c}{Data \& Rollout} \\
            Global Batch Size & 128 \\
            Max Prompt/Resp. & 2k / 8k \\
            Sampling Temp. $\tau$ & 1.0 \\
            Top-$p$ & 0.6 \\
            Rollout Group Size $G$ & 16 \\
            \midrule
            \rowcolor{gray!10} \multicolumn{2}{c}{RL Optimization} \\
            Learning Rate & $1 \times 10^{-6}$ \\
            Optimizer & AdamW \\
            Clip Ratio (L/H/C) & 0.2/0.28/10 \\
            KL Reference Loss & 0.0 \\
            Weight Decay & 0.01 \\
            \bottomrule
        \end{tabular}
        \end{sc}
    \end{minipage}
    \hfill
    \begin{minipage}{0.48\textwidth}
        \centering
        \caption{Hyperparameters for DPO.}
        \label{tab:dpo_config}
        \begin{sc}
        \begin{tabular}{lc}
            \toprule
            \textbf{Hyperparameter} & \textbf{Value} \\
            \midrule
            \rowcolor{gray!10} \multicolumn{2}{c}{Training Config} \\
            Learning Rate & $1e^{-6}$ \\
            $\beta$ (DPO temp.) & 0.05 \\
            Global Batch Size & 256 \\
            Grad. Accum. Steps & 4 \\
            Max Seq. Length & 4096 \\
            Training Epochs & 1 \\
            LR Scheduler & cosine \\
            Warmup Ratio & 0.1 \\
            Optimizer & AdamW \\
            Weight Decay & 0.01 \\
            \bottomrule
        \end{tabular}
        \end{sc}
    \end{minipage}
\end{table*}


\subsection{Ablation Study on Group Size and Sampling Temperature}
\label{sec:ablation}
To validate our hypothesis that large inter-group sampling enables models to perceive model’s upper-bound capability during training, we conduct ablation studies on two critical factors in our proposed balanced reward: rollout group size $G$ and sampling temperature $\tau$. 
Group size $G$ determines the diversity of reasoning trajectories available for preference optimization, while temperature $\tau$ modulates the stochasticity of action exploration. 
We train DAPO variants on DAPO-Math-17 with $G \in \{8, 16, 32\}$ and $\tau \in \{0.8, 1.0, 1.2\}$, evaluating tool-call efficiency and accuracy on the three benchmarks (i.e., \textbf{GSM8K}, \textbf{AIME24}, and \textbf{AIME25}).
The result is shown in~\cref{tab:ablation_group_temp}.

\begin{table*}[ht]
    \caption{Ablation study on group size ($G$) and sampling temperature ($\tau$), compared with baseline models. We report Accuracy (\%) and average Tool Calls (\#Calls) across three benchmarks. \textbf{Average} represents the arithmetic mean of accuracy and calls across all datasets.}
    \label{tab:ablation_group_temp}
    \centering
    \small
    \begin{sc}
        \begin{tabular}{cc | cc | cc | cc | cc}
            \toprule
             \multirow{2}{*}{$G$} & \multirow{2}{*}{$\tau$} & \multicolumn{2}{c|}{GSM8K} & \multicolumn{2}{c|}{AIME 24} & \multicolumn{2}{c|}{AIME 25} & \multicolumn{2}{c}{\textbf{Average}} \\
            \cmidrule(lr){3-4} \cmidrule(lr){5-6} \cmidrule(lr){7-8} \cmidrule(lr){9-10}
            & & Acc. $\uparrow$ & \#Calls $\downarrow$ & Acc. $\uparrow$ & \#Calls $\downarrow$ & Acc. $\uparrow$ & \#Calls $\downarrow$ & Acc. $\uparrow$ & \#Calls $\downarrow$ \\
            \midrule
            \multicolumn{2}{c|}{Qwen2.5-7B-inst} & 60.71 & 1.5 & 10.42 & 1.37 & 2.92 & 1.45 & 24.68 & 1.44 \\
            \multicolumn{2}{c|}{ReTool-7B} & 92.34 & 2.24 & 40.00 & 4.50 & 23.33 & 8.60 & 51.89 & 5.11 \\
            \midrule
             8  & 0.8 & 91.89 & 1.95 & 53.33 & 4.43 & 33.33 & 7.90 & 59.52 & 4.76 \\
             16 & 1.0 & 92.04 & 2.09 & 40.00 & 6.60 & 33.33 & 6.63 & 55.12 & 5.11 \\
             \textbf{32} & \textbf{1.2} & \textbf{92.34} & \textbf{2.30} & \textbf{50.00} & \textbf{5.40} & \textbf{30.00} & \textbf{6.47} & \textbf{57.45} & \textbf{4.72 } \\
            \bottomrule
        \end{tabular}
    \end{sc}
\end{table*}

\subsection{Evaluation Setting}
\label{sec:evaluation}
All evaluations are conducted with temperature $T=1.0$ and top-$p=1.0$, using vLLM \citep{kwon2023efficient} for efficient batched inference. 
For tool-augmented settings, we integrate a customized local sandboxed Python interpreter, where each tool call executes in an isolated environment. 
We report accuracy under $avg@8$ (average over 8 stochastic generations per prompt) to mitigate variance from non-deterministic decoding. 
For boundary analysis, we compute $avg@1024$ on the base model without tools to estimate intrinsic knowledge availability. 

\section{Mathematical Reasoning Benchmarks}
\label{sec:benchmark}
To evaluate the mathematical reasoning capabilities of Large Language Models (LLMs), we utilize several widely recognized benchmarks ranging from basic arithmetic to competition-level mathematics:

\begin{itemize}
    \item \textbf{GSM8K}: The Grade School Math 8K (including 7.5k train and 1.3k test) dataset consists of high-quality grade school math word problems \cite{cobbe2021training}. 
    It tests the model's ability to perform multi-step elementary reasoning.
    \item \textbf{AIME 24/25}: The American Invitational Mathematics Examination (2024 and 2025 versions) represents a significant leap in difficulty \cite{maa_aime}. 
    Each set 30-question short-answer tests are used to evaluate advanced reasoning and are currently a key discriminator for frontier models.
    We mix the two benchmarks in this paper.
\end{itemize}

\begin{table*}[th]
  \caption{  Based on avg@8 metric, statistical distribution of the number after division for each LLM.
  $\spadesuit$ denotes frontier models, $\heartsuit$ denotes RLVR-Trained models in TIR, $\clubsuit$ denotes OSS foundation models.
  }
  \label{tab:pre1_data}
  \centering
  \small
  \begin{sc}
    \begin{tabular}{lcccccc}
        \toprule
        \multirow{2}{*}{Model} & \multicolumn{2}{c}{GSM8K(1319)} & \multicolumn{2}{c}{AIME24/25 (60)} \\
        \cmidrule(lr){2-3} \cmidrule(lr){4-5}
              & Simple & Complex & Simple & Complex  \\
        \midrule
        $\spadesuit$ GPT-5.2       & 1260 & 59 & 50 & 4  \\
        $\spadesuit$ Gemini-3      & 1123 & 196 & 28 & 22  \\
        $\spadesuit$ Claude-4.5    & 360 & 959 & 22 & 37  \\
        $\spadesuit$ Deepseek-v3.2 & 1276 & 43 & 50 & 10  \\
        $\spadesuit$ Deepseek-R1   & 1137 & 182 & 13 & 38  \\
        $\spadesuit$ GLM-4.6   & 1205 & 114 & 15 & 45  \\
        \midrule
        $\heartsuit$ SimpleTIR-7B  & 1173 & 146 & 10 & 50 \\
        $\heartsuit$ ZeroTIR-7B    & 1227 & 92 & 4 & 56 \\
        $\heartsuit$ ReTool-7B     & 1195 & 124 & 1 & 59 \\
        $\heartsuit$ ReTool-32B    & 1270 & 49 & 10 & 50 \\
        \midrule
        $\clubsuit$ Qwen2.5-7B-inst  & 1234 & 85 & 9 & 51 \\
        $\clubsuit$ Qwen2.5-32B-inst & 1274 & 45 & 11 & 49 \\
        $\clubsuit$ Qwen3-8B         & 1274 & 45 & 33 & 27 \\
        $\clubsuit$ Llama3.1-7B-inst & 1180 & 139 & 5 & 55 \\ 
        $\clubsuit$ Llama3.2-3B-inst & 1092 & 227 & 3 & 57 \\ 
        \bottomrule
    \end{tabular}
  \end{sc}
  \vskip -0.1in
\end{table*}

\begin{table*}[th]
  \caption{Evaluation of reasoning capabilities and tool-integration efficiency on AIME 24/25 benchmark. 
  We report the average@8 accuracy and tool invocation frequency across simple and complex task samples. 
  $\spadesuit$ denotes frontier models, $\heartsuit$ denotes RLVR-Trained models in TIR, $\clubsuit$ denotes OSS foundation models.
  \textsuperscript{*} this model specially suppress invalid tool-use in RLVR training.
  }
  \label{tab:pre1_aime}
  \centering
  \small
  \begin{sc}
    \begin{tabular}{lcccccc}
        \toprule
        \multirow{2}{*}{Model} & \multicolumn{3}{c}{Simple Samples (\%)} & \multicolumn{3}{c}{Complex Samples (\%)} \\
        \cmidrule(lr){2-4} \cmidrule(lr){5-7}
              & avg@8 & avg@8 w/ tool & Tool Freq. & avg@8 & avg@8 w/ tool & Tool Freq. \\
        \midrule
        $\spadesuit$ GPT-5.2       & 99.75 & 52.25\down{47.50} & 1.33 & 0.00 & 0.00\up{0.00} & 0.47\down{0.86} \\
        $\spadesuit$ Gemini-3      & 88.16 & 71.05\down{17.11} & 1.92 & 23.86 & 39.20\up{15.34} & 3.76\up{1.84} \\
        $\spadesuit$ Claude-4.5    & 88.64 & 70.45\down{18.19} & 0.76 & 9.80 & 16.89\up{7.09} & 0.80\up{0.04} \\
        $\spadesuit$ Deepseek-v3.2 & 95.25 & 73.75\down{21.50} & 3.12 & 5.00 & 51.25\up{46.25} & 5.50\up{2.38} \\
        $\spadesuit$ Deepseek-R1   & 90.38 & 66.35\down{24.03} & 0.13 & 8.55 & 12.83\up{4.28} & 0.25\up{0.12} \\
        $\spadesuit$ GLM-4.6       & 85.00 & 69.17\down{15.83} & 0.07 & 12.78 & 16.39\up{3.61} & 0.11\up{0.04} \\
        \rowcolor{blue!10} $\spadesuit$ Average & 91.20 & 67.17\down{24.03} & 1.22 & 10.00 & 22.76\up{12.76} & 1.81\up{0.59} \\ 
        
        \midrule
        $\heartsuit$ SimpleTIR-7B\textsuperscript{*}  & 72.50 & 87.50\up{15.00} & 2.38 & 2.50 & 26.50\up{24.00} & 5.18\up{2.81} \\
        $\heartsuit$ ZeroTIR-7B    & 62.50 & 75.00\up{12.50} & 1.03 & 2.45 & 6.25\up{3.80} & 1.15\up{0.12} \\
        $\heartsuit$ ReTool-7B     & 56.93 & 87.50\up{30.57} & 1.12 & 2.84 & 17.65\up{14.81} & 1.63\up{0.51} \\
        $\heartsuit$ ReTool-32B    & 84.45 & 80.21\down{4.24} & 2.15 & 4.22 & 23.78\up{19.56} & 1.50\down{0.65} \\
        \rowcolor{blue!10} $\heartsuit$ Average & 69.10 & 82.55\up{13.46} & 1.67 & 3.00 & 18.55\up{15.54} & 2.37\up{0.70} \\
        \midrule
        $\clubsuit$ Qwen2.5-7B-inst  & 73.61 & 68.06\down{5.55} & 0.46 & 5.39 & 4.17\down{1.22} & 0.42\down{0.04} \\
        $\clubsuit$ Qwen2.5-32B-inst & 72.73 & 72.73\up{0.00} & 1.39 & 3.32 & 10.20\up{6.88} & 0.73\down{0.66} \\
        $\clubsuit$ Qwen3-8B         & 74.24 & 75.00\up{0.76} & 1.83 & 10.19 & 17.13\up{6.94} & 1.80\up{0.00} \\
        $\clubsuit$ Llama3.1-7B-inst & 80.00 & 57.50\down{22.50} & 1.00 & 0.91 & 3.18\up{2.27} & 0.95\down{0.05} \\ 
        $\clubsuit$ Llama3.2-3B-inst & 83.33 & 20.83\down{62.50} & 1.92 & 1.75 & 0.66\down{1.09} & 1.17\down{0.75} \\ 
        \rowcolor{blue!10} $\clubsuit$ Average & 76.78 & 58.82\down{17.96} & 1.32 & 4.31 & 7.07\up{2.76} & 1.01\down{0.31} \\
        
        \bottomrule
    \end{tabular}
  \end{sc}
  \vskip -0.1in
\end{table*}

\section{Additional Experimental Results and Details}

\subsection{For Quantifying Overuse in \cref{sec:quantify} }
\label{sec:add_res_overuse}
The supplementary results in \cref{tab:pre1_data} and \cref{tab:pre1_aime} provide a more granular view of tool-use efficiency across different reasoning difficulty levels. Several key observations emerge:

First, \cref{tab:pre1_data} reveals the stark contrast in task difficulty between GSM8K and AIME. While most models categorize the majority of GSM8K samples as \textit{simple}, the distribution shifts significantly on AIME 24/25, where foundation models like Llama3.2-3B categorize nearly all samples (57 out of 60) as \textit{complex}. 
Despite this, the performance penalty imposed by tools in simple samples remains prevalent. 
As shown in \cref{tab:pre1_aime}, frontier models ($\spadesuit$) suffer an average accuracy drop of $24.03\%$ on simple AIME samples when tools are integrated, echoing the interference effect observed in GSM8K. 
Notably, Llama3.2-3B exhibits the most extreme degradation, with its performance on simple samples plunging by $62.50\%$ upon tool use.
A small number of samples did not compute the final answer due to api access failures, so we removed these data.

Second, \cref{tab:pre1_data} reveals that RLVR-trained models ($\heartsuit$) identify the fewest simple samples compared to OSS models ($\clubsuit$), suggesting a heavy reliance on tools and a potential decline in intrinsic reasoning capabilities.
While ZeroTIR-7B and ReTool-7B show nominal performance gains on simple samples in \cref{tab:pre1_aime}, these results are statistically insignificant due to the extremely small sample sizes ($n=4$ and $n=1$, respectively).
In contrast, SimpleTIR-7B* achieves robust gains across both simple (+15.00\%) and complex (+24.00\%) subsets, demonstrating that the feedback masking mechanism effectively suppresses invalid tool-use and optimizes reasoning efficiency.


\subsection{Knowledge–Behavior Analysis in other models For~\cref{sec:boundary}}
\label{sec:add_res_boundary}
\begin{figure*}[ht]
  \vskip 0.2in
  \begin{center}
    \begin{subfigure}[t]{0.9\linewidth}
        \includegraphics[width=0.99\linewidth]{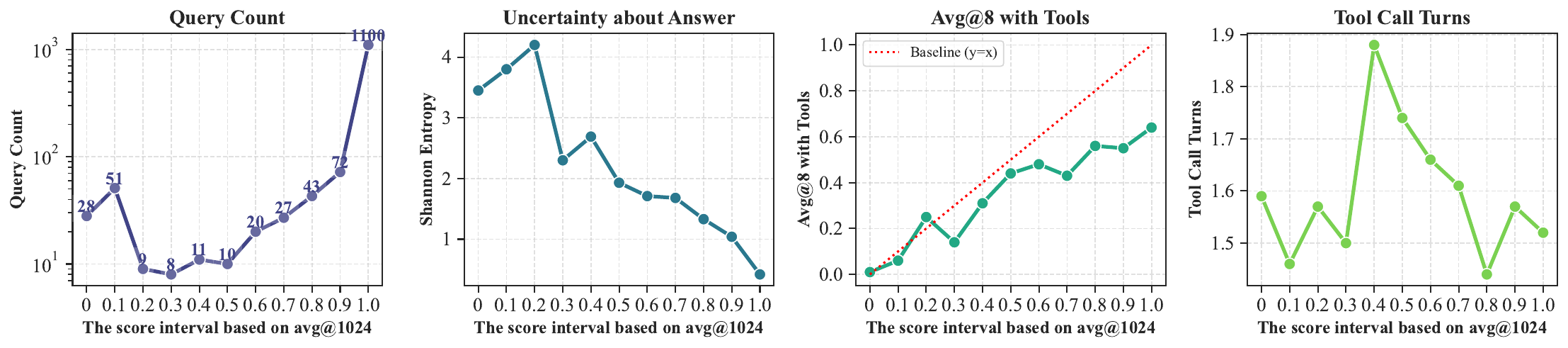}
        \caption{Qwen2.5-7B-Instruct.}
        \label{fig:qwen2.5}
    \end{subfigure}
    \begin{subfigure}[t]{0.9\linewidth}
        \includegraphics[width=0.99\linewidth]{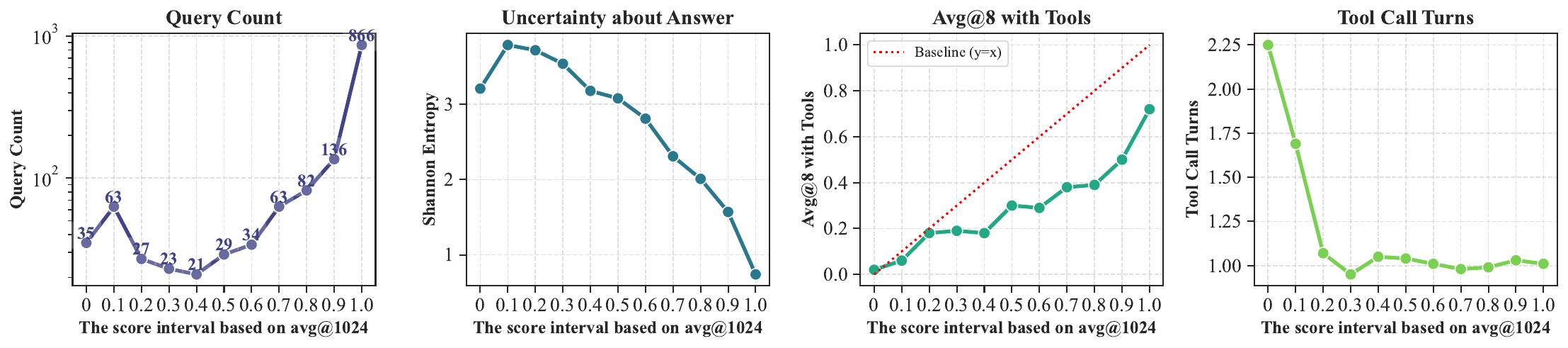}
        \caption{Llama-3.1-8B.}
        \label{fig:llama3.1}
    \end{subfigure}
        \begin{subfigure}[t]{0.9\linewidth}
        \includegraphics[width=0.99\linewidth]{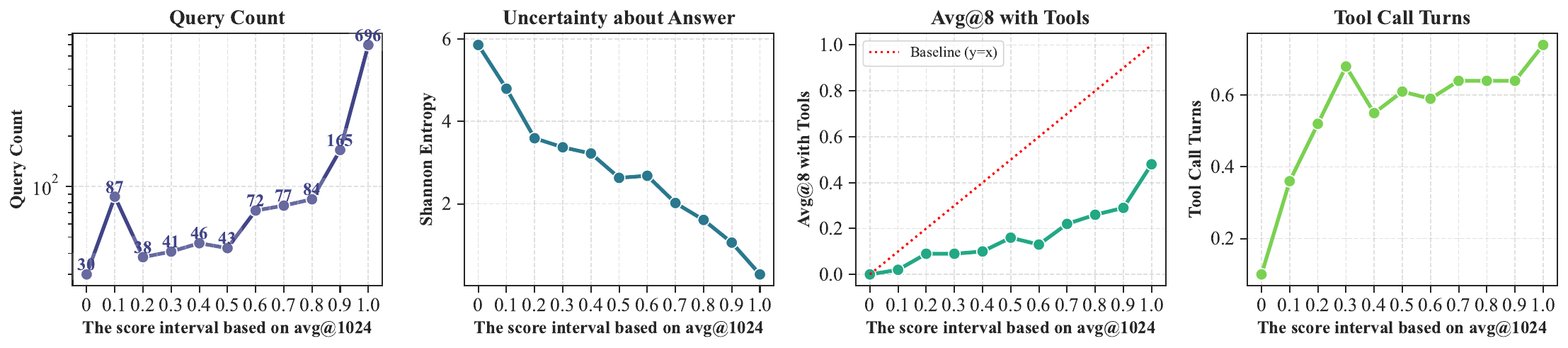}
        \caption{Llama-3.2-3B.}
        \label{fig:llama3.2}
    \end{subfigure}
  \hfill
    \begin{subfigure}[t]{0.9\linewidth}
        \includegraphics[width=0.99\linewidth]{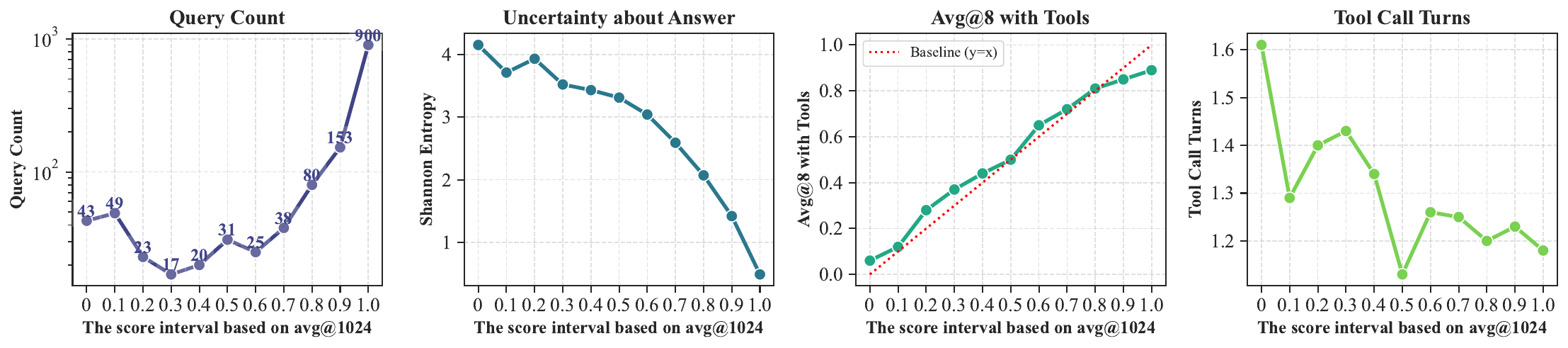}
        \caption{ReTool-7B.}
        \label{fig:retool-7b}
    \end{subfigure}
    \begin{subfigure}[t]{0.9\linewidth}
        \includegraphics[width=0.99\linewidth]{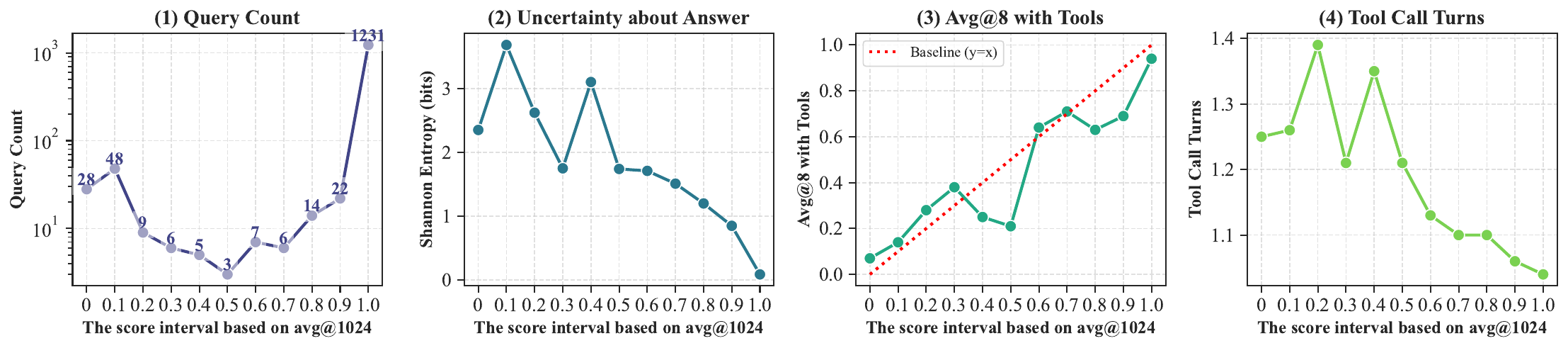}
        \caption{Qwen2.5-32B-Instruct.}
        \label{fig:qwen3-32b}
    \end{subfigure}
    \caption{
      Tool-use behavior and performance across the model’s epistemic boundary for other models.
      An anomalous dip in Qwen2.5-32B-Instruct's tool-augmented curve within $avg@1024 \in (0.4, 0.5)$ (Figure~\ref{fig:qwen3-32b}) stems from limited sample counts ($n\leq5$) in this bin, highlighting the importance of sufficient data density for stable boundary estimation.
    }
    \label{fig:boundary2}
  \end{center}
\end{figure*}

To assess the generality of the knowledge epistemic illusion, we extend our analysis to five additional models: Qwen2.5-7B-Instruct, Qwen2.5-32B-Instruct, ReTool-7B, Llama-3.1-8B, and Llama-3.2-3B. Results are presented in Figure~\ref{fig:boundary2}.

We identify two distinct behavioral patterns. First, Qwen2.5-7B-Instruct, Llama-3.1-8B, and Llama-3.2-3B display \textit{indiscriminate tool use}: tool augmentation consistently reduces $avg@8$ accuracy across all knowledge regimes. 
This suggests these models lack the basic ability to use tools.
Second, Qwen2.5-32B-Instruct shows: in low-to-moderate knowledge regimes ($avg@1024 \in [0, 0.7]$), tool augmentation improves $avg@8$ accuracy, indicating effective compensation for knowledge gaps. 
However, in high-knowledge regions ($avg@1024 > 0.7$), tool use degrades performance, revealing residual overuse despite superior capability.

\subsection{For Training visualization in ~\cref{sec:training}}
\label{sec:add_res_training}
\begin{figure}[th]
  \vskip 0.1in
  \begin{center}
    \centerline{\includegraphics[width=0.9\columnwidth]{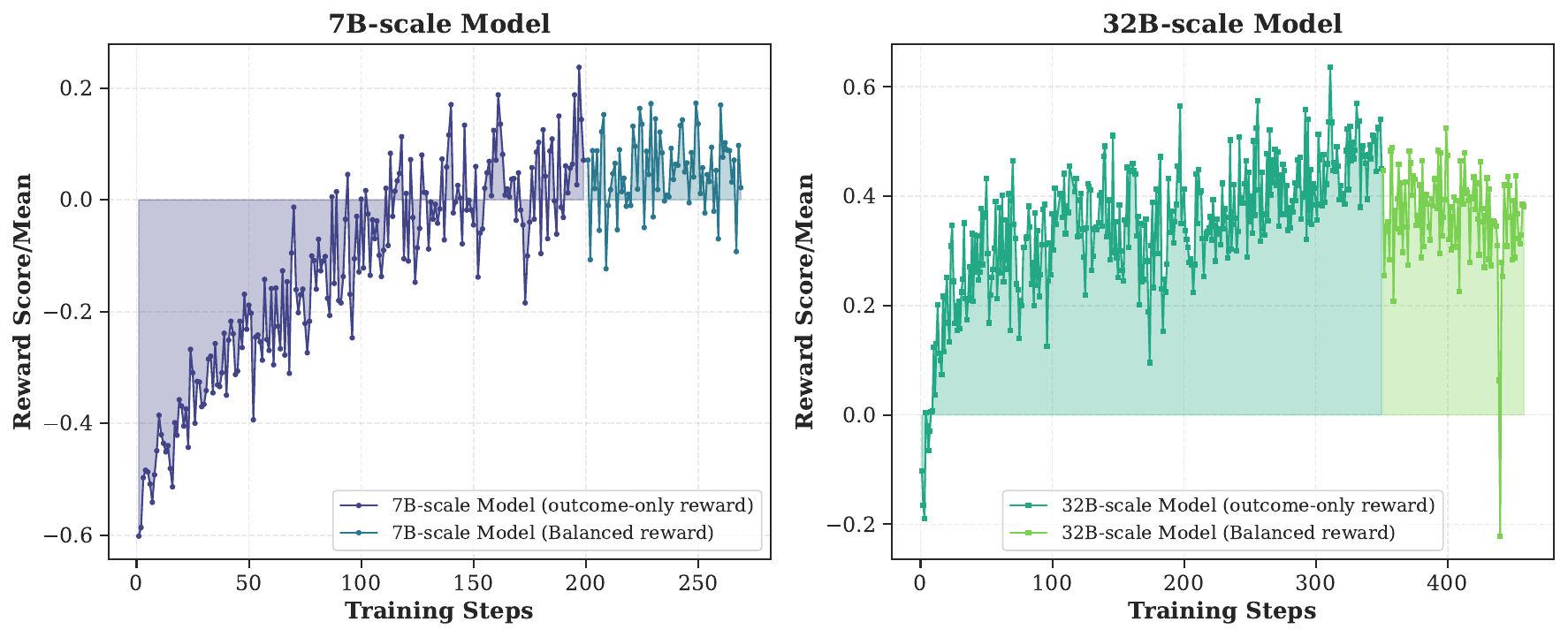}}
    \caption{
      Comparison of reward trajectories during the training process for Qwen-7B (left) and Qwen-32B (right). The solid lines represent the initial training phase, while the dashed lines denote the continual training stage. 
    }
    \label{fig:training_qwen32}
  \end{center}
  \vskip -0.2in
\end{figure}

To evaluate the stability of the transition between different training stages and model scales, we visualize the trajectory of the reward scores for both Qwen-7B and Qwen-32B models. As illustrated in \cref{fig:training_qwen32}, the training process is divided into two phases: an initial RLVR (e.g., DAPO algorithm) training with outcome-only reward  phase, followed by our balanced continual training stage.

A key observation is that the {reward scores exhibit remarkable continuity} when transitioning to the second stage. 
Despite the shift in training objectives or model configurations, there is no significant degradation in the reward signals, suggesting that the model effectively preserves learned representations
This finding shows our {balanced outcome-efficiency reward within the RLVR framework} significantly alleviates the \textit{tool overuse} problem while maintaining robust training stability.
This stability is crucial for ensuring the convergence of Reinforcement Learning (RL) algorithms, as it provides a consistent baseline for advantage estimation.
By penalizing redundant tool use without sacrificing final accuracy, the model learns to generate more concise and purposeful tool calls.

\begin{figure}[th]
  \vskip 0.1in
  \begin{center}
    \begin{subfigure}[t]{0.49\linewidth}
        \centering
        \includegraphics[width=0.99\linewidth]{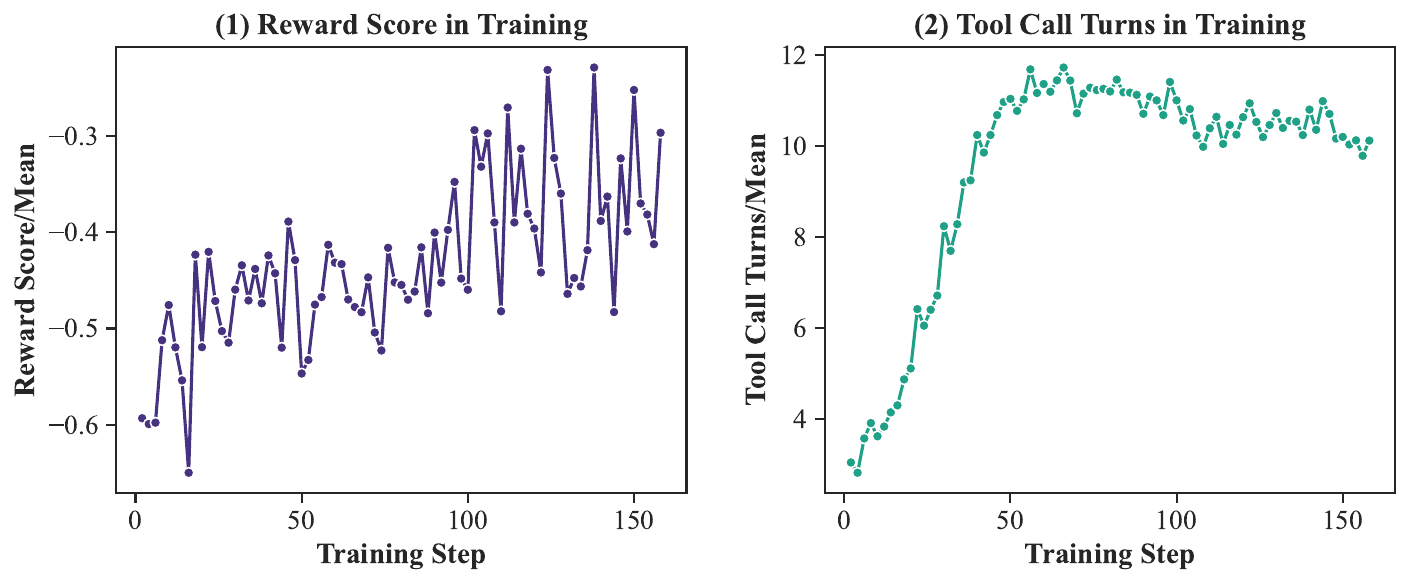}
        \caption{Qwen2.5-3B-Instruct}
        \label{fig:qwen_3b}
    \end{subfigure}
    \hfill
    \begin{subfigure}[t]{0.49\linewidth}
        \centering
        \includegraphics[width=0.99\linewidth]{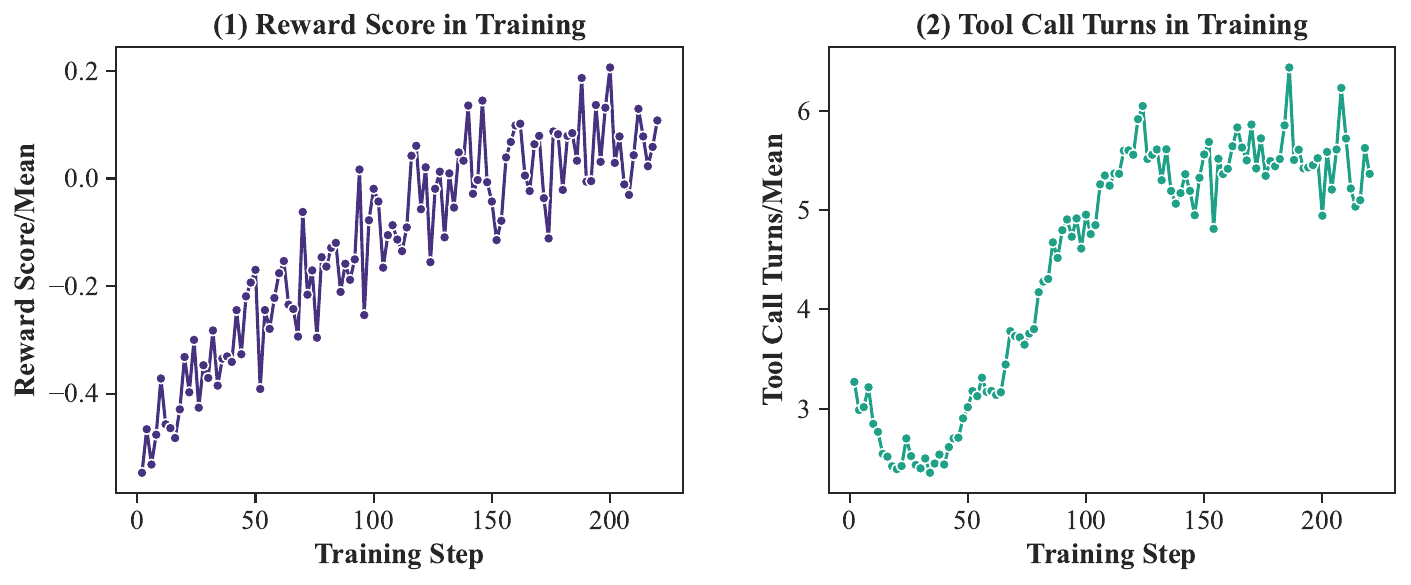}
        \caption{Qwen2.5-7B-Instruct}
        \label{fig:qwen_7b}
    \end{subfigure}
    
    \vskip 0.1in
    \begin{subfigure}[t]{0.49\linewidth}
        \centering
        \includegraphics[width=0.99\linewidth]{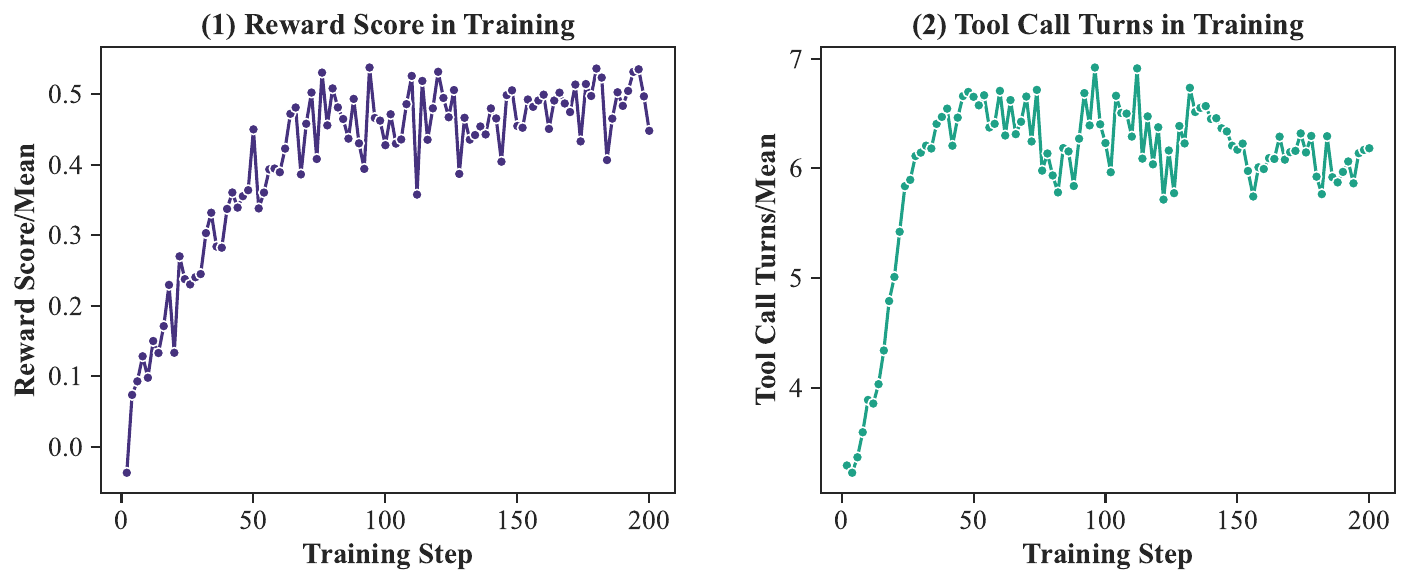}
        \caption{Qwen2.5-32B-Instruct}
        \label{fig:qwen_32b}
    \end{subfigure}
    \caption{
      Visualization of mean rewards and tool call turns during the training of Qwen2.5-3B/7B/32B-Instruct models. 
    }
    \label{fig:training_qwen_series}
  \end{center}
  \vskip -0.1in
\end{figure}

Besides, we extend the visualization in reproducing the training process on the Qwen2.5-Instruct series, including the 3B, 7B, and 32B variants, strictly following the ReTool~\cite{feng2025retool} methodology. 
The following visualizations (\cref{fig:training_qwen_series}) illustrate these reward dynamics across different model scales under a consistent set of hyperparameters.

\end{document}